\documentclass[letterpaper]{article} 
\usepackage{aaai2026}  
\usepackage{times}  
\usepackage{helvet}  
\usepackage{courier}  
\usepackage[hyphens]{url}  
\usepackage{graphicx} 
\usepackage{natbib}  
\usepackage{caption}
\usepackage{CJKutf8}
\usepackage{algorithm}
\usepackage{algorithmic}
\usepackage{booktabs}    
\usepackage{multirow}    
\usepackage{array}       
\usepackage{caption}
\usepackage{xcolor}
\usepackage{float}
\usepackage{subcaption}
\newcolumntype{P}[1]{>{\RaggedRight\arraybackslash}p{#1}}

\usepackage{xcolor}

\usepackage{newfloat}
\usepackage{listings}
\DeclareCaptionStyle{ruled}{labelfont=normalfont,labelsep=colon,strut=off} 
\floatstyle{ruled}
\newfloat{listing}{tb}{lst}{}
\floatname{listing}{Listing}
\nocopyright

\usepackage{times}
\usepackage{latexsym}
\usepackage{adjustbox}
\usepackage{graphicx}
\usepackage{xcolor}
\usepackage{makecell}
\usepackage{algorithm}
\usepackage{algorithmic}
\usepackage{multirow}
\usepackage{float}
\usepackage{tabularx}
\usepackage{booktabs}
\usepackage[a4paper, margin=1in]{geometry}
\usepackage{array}       
\usepackage{ragged2e}    
\usepackage[T1]{fontenc}
\usepackage{CJKutf8}
\usepackage[utf8]{inputenc}
\usepackage{microtype}
\usepackage{inconsolata}
\usepackage{amsmath}
\usepackage{graphicx}

\title{AdaMCoT: Rethinking Cross-Lingual Factual Reasoning through \\Adaptive Multilingual Chain-of-Thought}

\author{
    Weihua Zheng\textsuperscript{\rm 1,\rm 2,*}, 
    Xin Huang\textsuperscript{\rm 1,*},
    Zhengyuan Liu\textsuperscript{\rm 1}, 
    Tarun Kumar Vangani\textsuperscript{\rm 1},\newline
    Bowei Zou\textsuperscript{\rm 1},
    Xiyan Tao\textsuperscript{\rm 1},
    Yuhao Wu\textsuperscript{\rm 2},
    Ai Ti Aw\textsuperscript{\rm 1},
    Nancy F. Chen\textsuperscript{\rm 1},
    Roy Ka-Wei Lee\textsuperscript{\rm 2}
}
\affiliations{
    \textsuperscript{\rm 1}Institute for Infocomm Research, A*STAR, Singapore\\
    \textsuperscript{\rm 2}Singapore University of Technology and Design\\
    {zheng\_weihua@a-star.edu.sg, liu\_zhengyuan@a-star.edu.sg, \\nfychen@a-star.edu.sg, roy\_lee@sutd.edu.sg}
}

\begin{document}

\maketitle

\begin{abstract}
Large language models (LLMs) have shown impressive multilingual capabilities through pretraining on diverse corpora. Although these models show strong reasoning abilities, their performance varies significantly between languages due to the imbalanced distribution of training data. Existing approaches using sample-level translation for extensive multilingual pretraining and cross-lingual tuning face scalability challenges and often fail to capture nuanced reasoning processes across languages.
In this paper, we introduce \texttt{AdaMCOT} (Adaptive Multilingual Chain-of-Thought), a framework that enhances multilingual factual reasoning by dynamically routing thought processes in intermediary ``thinking languages'' before generating target-language responses. \texttt{AdaMCOT} leverages a language-agnostic core and incorporates an adaptive, reward-based mechanism for selecting optimal reasoning pathways without requiring additional pretraining. Our comprehensive evaluation across multiple benchmarks demonstrates substantial improvements in both factual reasoning quality and cross-lingual consistency, with particularly strong performance gains in low-resource language settings. An in-depth analysis of the model’s hidden states and semantic space further elucidates the underlying mechanism of our method. The results suggest that adaptive reasoning paths can effectively bridge the performance gap between high and low-resource languages while maintaining cultural and linguistic nuances.

\textbf{Code}: \url{https://github.com/Social-AI-Studio/AdaMCoT}
\end{abstract}

\section{Introduction}

Large Language Models (LLMs) exhibit strong reasoning capabilities but demonstrate significant performance disparities across languages, favoring major languages like English \cite{achiam2023gpt,dubey2024llama}. This linguistic bias limits accessibility for diverse global communities \cite{singh-etal-2024-aya}, while translation-based solutions prove inadequate for robust multilingual reasoning due to artifacts and failure to capture cross-linguistic logical nuances.
Current approaches to enhance multilingual LLMs operate at two levels: data-level methods that utilize large-scale multilingual corpora for continual pretraining \cite{xuparadigm,yang2023bigtranslate}, instruction fine-tuning \cite{li2023bactrian,zhang-etal-2024-enhancing-multilingual}, and cross-lingual alignment \cite{zhang2023bayling,zhu2023extrapolating}; and representation-level methods that align embeddings across languages \cite{li-etal-2024-improving-context,li2024prealign,tang2022align}. However, these approaches typically demand extensive training data and fail to deliver consistent reasoning improvements \cite{zhu2023extrapolating,li-etal-2024-improving-context}.

Recent studies reveal that LLMs possess a language-agnostic reasoning core enabling cross-lingual reasoning \cite{tang-etal-2024-language,zhao2024large,schut2025multilingualllmsthinkenglish,wang2025lostmultilingualitydissectingcrosslingual,fierro2025multilinguallanguagemodelsremember}. Nevertheless, certain factual knowledge remains language-dependent due to imbalanced training distributions and regional linguistic features. Despite robust reasoning cores in major languages, LLMs struggle to effectively integrate cross-lingual factual knowledge into reasoning processes, particularly for low-resource languages with limited training exposure, necessitating frameworks that leverage language-specific strengths while preserving cultural context.

In this work, we present \textbf{\texttt{AdaMCOT}}, a framework that enhances multilingual factual reasoning by bridging the gap between LLMs’ language-agnostic reasoning and language-dependent factual knowledge. Our core insight is that different languages offer unique advantages—due to structural features, cultural grounding, or training representation—that can improve reasoning performance. For example, languages rich in logical connectives may support deductive reasoning, while those with strong mathematical vocabulary may better facilitate quantitative tasks. \texttt{AdaMCOT} exploits these strengths by routing reasoning through intermediate ``thinking languages'' before generating the final response in the target language.

The framework is built on two principles: (1) \textit{Dynamic Routing Optimization}, which learns to select optimal intermediate languages based on task characteristics and prior performance; and (2) \textit{Cross-Lingual Knowledge Integration}, which synthesizes insights across languages to enhance output robustness. \texttt{AdaMCOT} uses a reward-based mechanism to evaluate candidate reasoning paths, optimizing for answer accuracy, consistency, and fluency. This enables adaptive, efficient multilingual reasoning without requiring additional multilingual pretraining.

Our experiments on multilingual TruthfulQA, CrossAlpaca-Eval 2.0, Cross-MMLU, and Cross-LogiQA show that \texttt{AdaMCOT} significantly enhances cross-lingual reasoning and consistency. Low-resource languages particularly benefit when reasoning is routed through linguistically related, high-resource counterparts, highlighting the transferability of reasoning patterns within language families. The adaptive routing mechanism also learns task-sensitive heuristics, favoring technically precise languages for scientific queries and expressive ones for sentiment tasks. To understand these gains, we use Logit Lens \cite{nostalgebraist2020logitlens} and UMAP \cite{mcinnes2020umapuniformmanifoldapproximation} to visualize the reasoning space. These analyses reveal that optimal routing enhances performance and aligns semantic spaces across languages more tightly.

Overall, our findings validate \texttt{AdaMCOT}’s effectiveness and shed light on multilingual reasoning dynamics in LLMs. The observed benefits of language-specific routing suggest that different languages encode distinct cognitive priors, offering valuable insights for more robust, nuanced reasoning.

\section{Related Works}
LLMs exhibit multilingual capabilities through pretraining on diverse corpora \cite{bang2023multitask,qin2025survey}. While primarily developed for resource-rich languages such as English, French, and Chinese, LLMs have shown unexpected proficiency across languages \cite{dubey2024llama}. However, uneven training data distribution at language level leads to performance disparities between high and low-resource languages \cite{qi-etal-2023-cross,wang-etal-2024-seaeval,chua2024crosslingual}.

\begin{figure*}[t]
\centering
\includegraphics[trim=0 10 5 0, clip, width=0.90\linewidth]{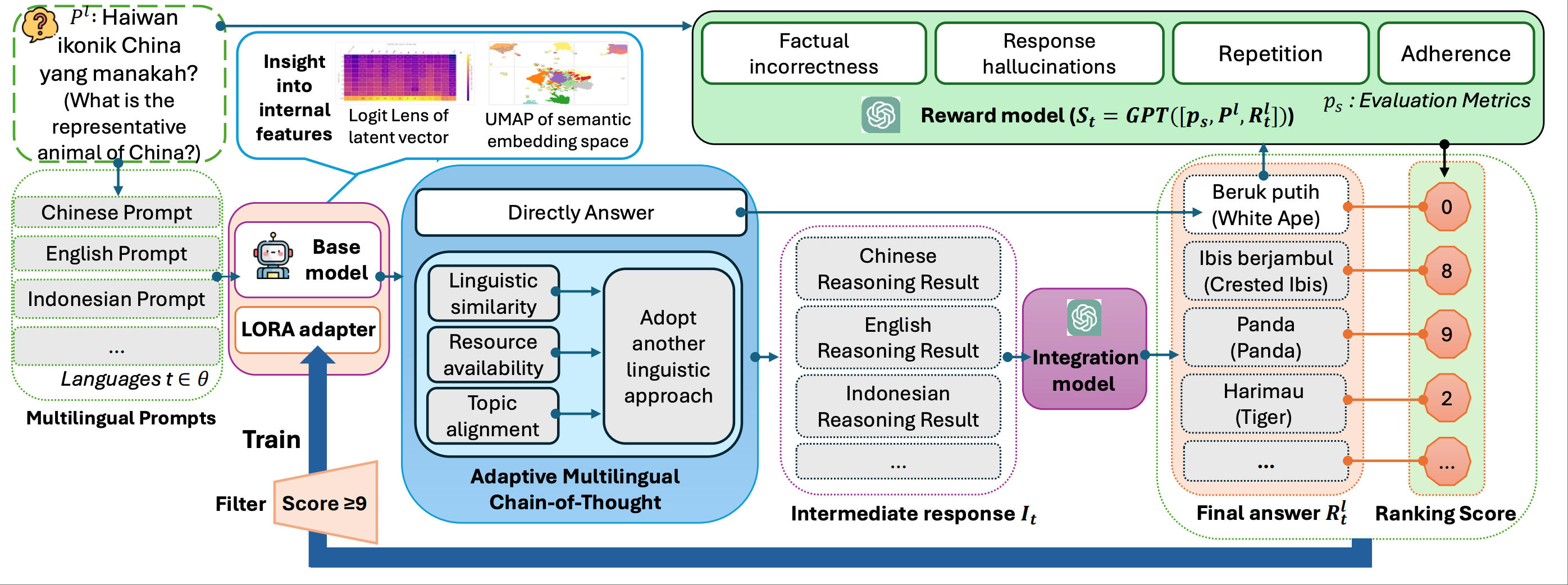}
\caption{Overview of the \texttt{AdaMCOT} framework. The input in the example shown in the figure is a question in Malay.}\label{fig:prefdata}
\vspace{-0.1cm}
\end{figure*}

Recent studies on language processing dynamics show that LLMs often adopt an internal pivot language—typically their primary training language (e.g., English) \cite{wendler2024llamas,zhong2024beyond}. Several prompting-based methods have attempted to harness this property for cross-lingual reasoning. XLT \cite{huang2023not} and Cross-lingual Prompting \cite{qin2023cross} rely on fixed reasoning paths via templates, while AutoCAP \cite{zhang2024autocap} uses automatic alignment planning for zero-shot chain-of-thought reasoning. However, AutoCAP's multi-language voting mechanism can introduce noise, especially when incorrect reasoning paths dominate. This is exacerbated in open-source models with uneven reasoning capabilities across languages. Moreover, the need for multiple generations leads to substantial inference overhead.

Prompt-only methods also fall short: without model fine-tuning, LLMs tend to default to the prompt language for reasoning. In contrast, our method preserves inference efficiency and overcomes language bias through instruction tuning, enabling more accurate and diverse language routing. Mixed-language strategies, such as xCoT \cite{chai2025xcot}, combine translated inputs and prompts across languages \cite{zhu2024question}, but risk semantic drift and error propagation due to fragmented translations.

Our approach instead preserves the original query language and adaptively selects an auxiliary “thinking language” for the reasoning phase. This decoupling enables the model to leverage language-specific strengths, while mitigating translation artifacts. The result is a more flexible, efficient, and context-aware multilingual reasoning process. Additional related work is discussed in Appendix A.1.

\section{Methodology}

We propose a novel framework, \texttt{AdaMCOT}, for enhancing multilingual reasoning in LLMs through \emph{adaptive chain-of-thought} prompting. The key idea is to improve the factual reasoning performance in a target language by routing intermediate reasoning steps through one or more auxiliary “thinking languages.” While LLMs possess a language-agnostic reasoning core, their outputs often vary across languages due to differences in training data coverage, linguistic structure, and representation alignment. An overview of the \texttt{AdaMCOT} framework is illustrated in Figure~\ref{fig:prefdata}, which depicts the generation of intermediate reasoning steps, reward-based selection, and training flow.

\texttt{AdaMCOT} dynamically selects auxiliary languages that are linguistically similar to the input, rich in relevant knowledge, or well-aligned with the model’s internal representations. These languages are used to guide the model’s internal reasoning process before producing a response in the original language. For instance, English, which is often overrepresented in training data, can serve as an effective intermediary for reasoning tasks posed in underrepresented languages.

To facilitate this, we introduce a \textit{dual-pathway mechanism}: (i) \textit{Cross-Lingual Chain-of-Thought}, which performs intermediate reasoning in one or more auxiliary languages before generating the final answer in the target language. (ii) \textit{Direct Generation}, which bypasses cross-lingual reasoning and directly produces the response in the same language as the prompt, useful for well-supported languages or linguistically sensitive tasks.

The model learns to select between these pathways using a reward-based fine-tuning approach. A strong LLM-based reward model (e.g., GPT-4o) evaluates the quality of generated outputs and provides feedback based on multiple criteria such as factual correctness, fluency, and instruction adherence. This reward signal is used to fine-tune the base model, enabling it to dynamically adapt its reasoning strategy to the input context.

\subsection{Candidate Response Generation}
\label{para:Candidate Response Generation}

The pre-training of multilingual LLMs typically relies on monolingual next-token prediction, which makes these models more proficient at generating responses in the same language as the input prompt. As a result, reasoning and generation are naturally biased toward the prompt language. To enable reasoning in a different language from the target language, we adopt two strategies for generating a diverse pool of candidate outputs, given a prompt $P_l$ in language $l$:

\begin{itemize}
\item \textbf{Cross-Lingual Chain-of-Thought}. This approach uses intermediate reasoning steps in auxiliary languages to guide the creation of the final response. We begin by rephrasing the original prompt into auxiliary languages such as English, Chinese, and Indonesian, then instruct the base LLM to generate responses for each linguistic variant. As the internal knowledge of the model may vary between languages, these responses may differ and reflect unique linguistic and cultural nuances. To ensure that the final answer maintains linguistic consistency with the input instruction while preserving the diverse knowledge embedded in the intermediate reasoning processes, we introduce an integration model(GPT-4o) to transform the intermediate reasoning into a final answer in the target language. Formally, given a prompt $P_l$ in language $l$ and an intermediate response $I_t$ in auxiliary language $t$, we instruct the integration model to produce a final response $R_l^t$ in language $l$, where $R_l^t$ keeps the semantic meaning of $I_t$ while adhering to the original instruction $P_l$.

\item \textbf{Direct Generation}. Direct Generation creates responses in the intended language without relying on any auxiliary languages. Given a prompt $P^l$ in language $l$, the model directly produces the response $R_l^l$ in the same language. This strategy is especially useful if the model is already strong in that language or in situations where using multiple languages might degrade the performance, such as linguistic dependent tasks like writing poetry.
\end{itemize}

With these two strategies, \texttt{AdaMCOT} generates diverse responses across multiple language pathways, which can then be ranked by a powerful LLM. Because certain knowledge may not be available or shared in all languages, the approach helps reduce the risk of factual hallucinations.

\subsection{Candidate Response Ranking}

We leverage a strong LLM, acting as a reward model, to score responses generated via different reasoning pathways (including direct generation) and select the optimal one based on the scores. Given an input prompt $P_l$ in language $l$ and a set of candidate auxiliary languages $\theta$, a final response $R_l^t$ is produced in the target language $l$ by employing a language $t \in \theta$ for intermediate reasoning. The reward model then provides a text-based score for each generated response, whether produced through an auxiliary language or via direct generation.

We leverage GPT-4o as both the reward model and the integration LLM. A detailed rationale for the selection of the reward model, risk discussion, along with human preference consistency evaluation, is provided in Appendix A.2. To evaluate response quality, a specialized prompt \(p_s\) guides the model to jointly assess four metrics: \textit{factual accuracy}, \textit{hallucination avoidance}, \textit{redundancy}, and \textit{instruction compliance}, producing a composite score \(S_t\) on a Likert-like scale 0-10, as formalized in Equation \ref{eq:1}.
\begin{equation}
S_t = GPT([p_s, P_l, R_l^t])
\label{eq:1}
\end{equation}

\subsection{\texttt{AdaMCOT} Fine-Tuning}

We fine-tune the base model using only high-quality outputs, specifically, those with reward scores \( S_t \geq 9 \). For each training instance, we select the highest-scoring candidate response as the optimal reasoning pathway. Both the intermediate reasoning \( I_t \) and the final response \( R_l^t \) are generated by the model itself, not derived from human-annotated datasets. This self-supervised setup mitigates the risk of knowledge forgetting during fine-tuning.

To incorporate reasoning structure, training examples are formatted with special prompts. For a given input query \( P_l \) in language \( l \), intermediate reasoning \( I_t \) in auxiliary language \( t \), and final response \( R_l^t \) in language \( l \), we construct the training sequence as: $P_l\ [P_l^q]\ I_t\ [P_l^r]\ R_l^t$
where \( P_l^q \) marks the beginning of the reasoning phase, and \( P_l^r \) indicates the transition to the final answer. Importantly, the final response \( R_l^t \) is not a summary of \( I_t \), but is generated \emph{conditioned on} it. This separation ensures the model can produce fluent, instruction-aligned outputs in the target language, while flexibly leveraging reasoning in another language.

For direct generation (i.e., no intermediate reasoning), the format is simplified to:$P_l\ [P_l^q]\ R_l^t$. During fine-tuning, we apply an attention mask to the input prompt \( P_l \), setting its attention weights to zero. This ensures the model focuses on learning to predict the correct reasoning path and final answer, rather than memorizing prompt surface forms.

\begin{figure*}[t]
  \centering
  \includegraphics[trim=0 10 0 8, clip, width=0.85\textwidth]{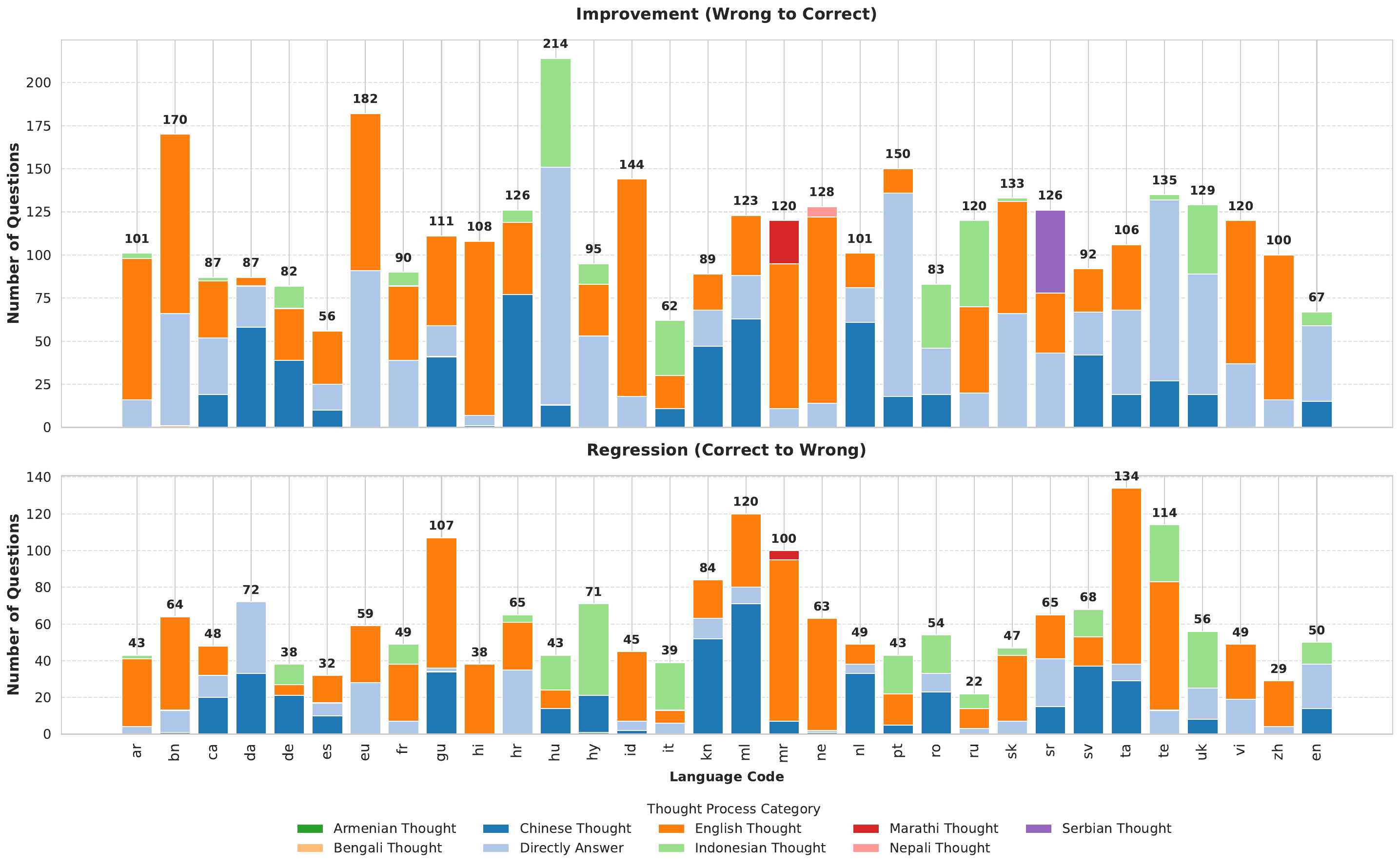}
  \caption{Distribution of Reasoning Pathway Selections on the mTruth Dataset: LLaMA3.1-8B-\texttt{AdaMCOT} vs. Base Model.} 
  \label{fig:changes}
\end{figure*}

\begin{figure*}[ht] 
  \centering

  \begin{subfigure}{0.49\textwidth}
    \centering
    \includegraphics[trim=0 130 0 110, clip, width=\textwidth]{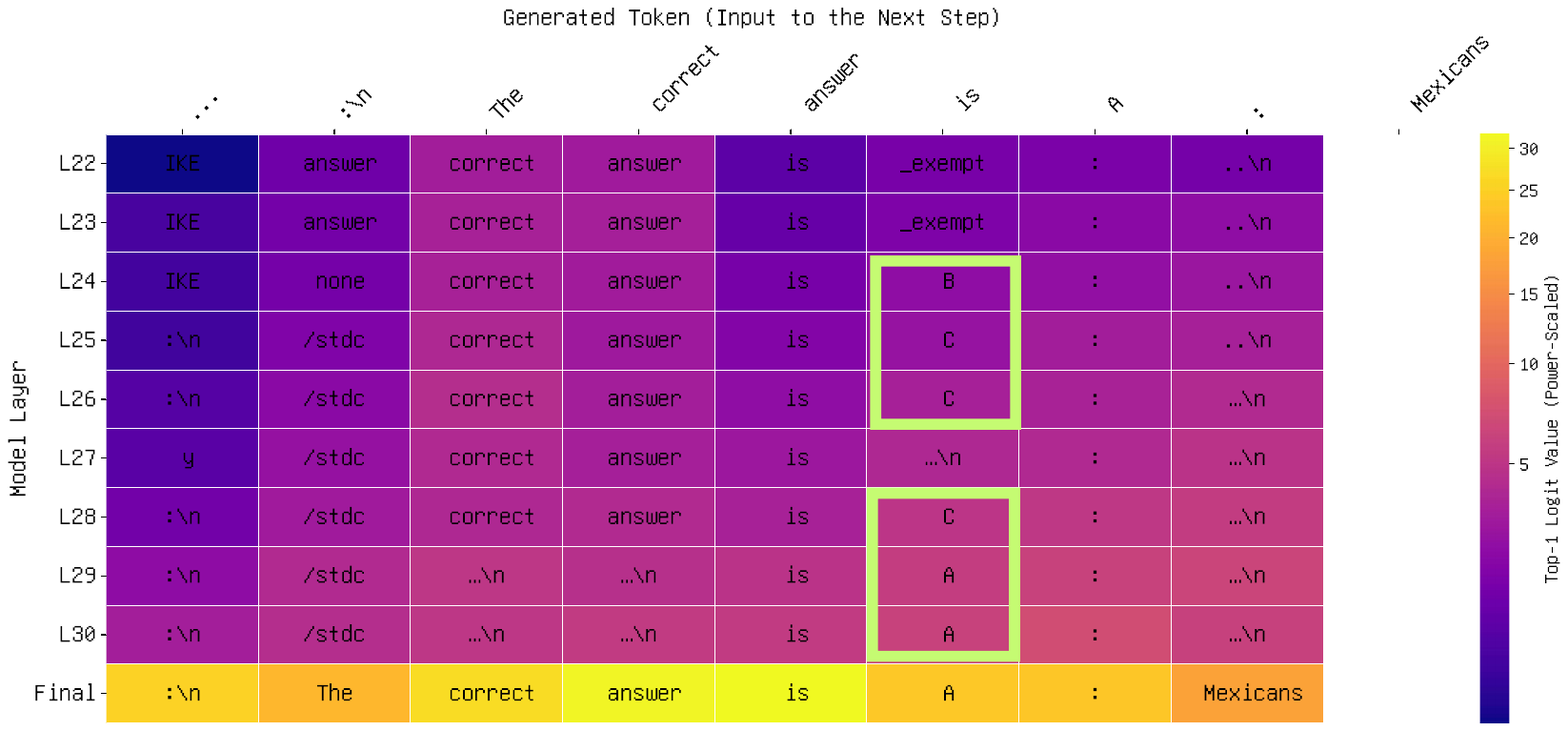}
    \subcaption{Latent space for token 8–15}
    \label{fig:llama-adop-hot-15}
  \end{subfigure}
  \hfill 
  \begin{subfigure}{0.49\textwidth}
    \centering
    \includegraphics[trim=0 130 0 110, clip, width=\textwidth]{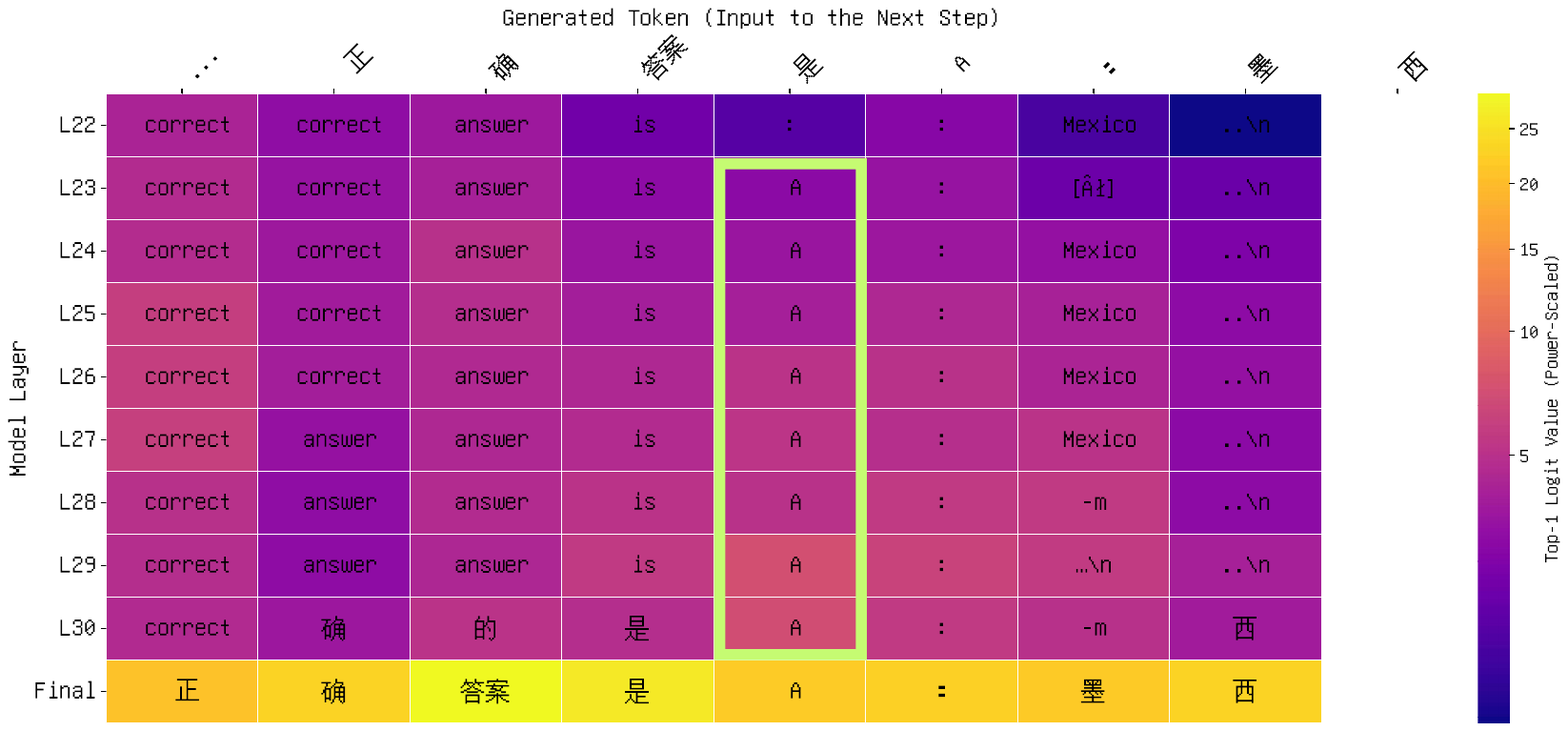}
    \subcaption{Latent space for token 83–90}
    \label{fig:llama-adop-hot-88-103}
  \end{subfigure}

  \vskip\baselineskip 

  \begin{subfigure}{1.0\textwidth} 
    \centering
    \includegraphics[trim=0 390 0 28, clip, width=\textwidth]{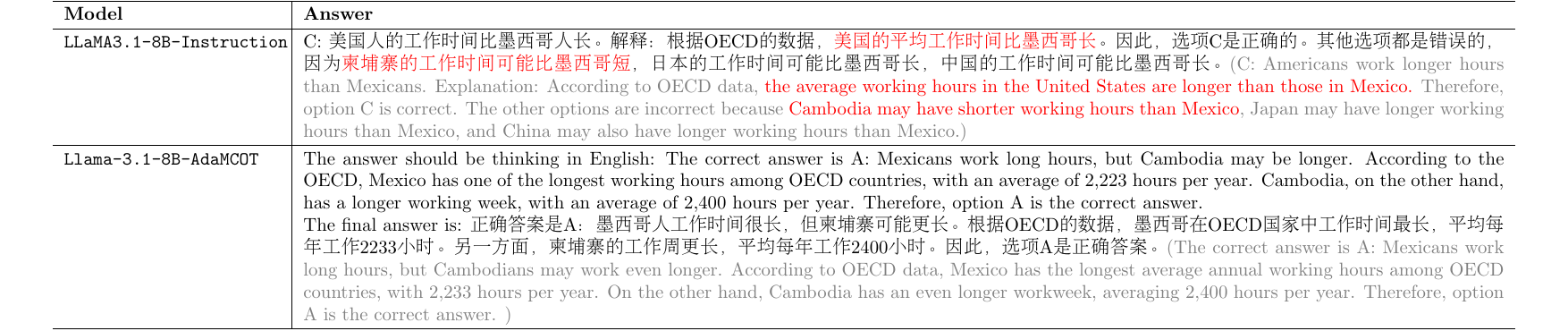}
    \subcaption{Answers of Llama-3.1-8B-Instruction and Llama-3.1-8B-\texttt{AdaMCOT}}
    \label{fig:answer}
  \end{subfigure}

  \caption{Logit lens maps for Llama-3.1-8B-\texttt{AdaMCOT} (a and b) and answers of Llama-3.1-8B-Instruction and Llama-3.1-8B-\texttt{AdaMCOT} (c), when prompted with \textit{``Which countries have longer working hours than Mexico? A.Mexicans work long hours, but Cambodians might work even longer ... F: Mexicans work fewer hours than the Chinese."} in Chinese. (a) and (b) highlight key changes in the model's intermediate layers with green boxes. Model hallucinations are highlighted in red in (c).}
  \label{fig:main-figure}

\end{figure*}

\section{Experimentation}

\subsection{Experiment Setup}

\subsubsection{Base Models.}
We apply \texttt{AdaMCOT} to two strong multilingual open-source LLMs: LLaMA3.1-8B-Instruct \cite{dubey2024llama} and Qwen2.5-7B-Instruct \cite{qwen2025qwen25technicalreport}.

\subsubsection{Primary Thinking Languages.}
We select English, Chinese, and Indonesian as intermediary reasoning languages. English and Chinese offer broad but distinct knowledge coverage, while Indonesian (a low-resource language) tests our hypothesis that domain-specific reasoning may benefit from culturally aligned languages.

\subsubsection{Training Datasets.}
To train the model for optimal reasoning path selection, we compile diverse instruction data: 1M English prompts from OpenHermes 2.5 \cite{OpenHermes} and 1.1M Chinese prompts from Firefly \cite{Firefly}. We augment this with multilingual prompts in English, Chinese, and Indonesian generated by GPT-4o. The final fine-tuning dataset is constructed using the candidate response generation and ranking process detailed in Section~\ref{para:Candidate Response Generation}. Full training details are in Appendix A.6.

\subsection{Evaluation Datasets}

\newsavebox{\myfirsttablebox}

\begin{table*}[h!]
\centering
\small

\sbox{\myfirsttablebox}{%
\begin{tabular}{l|ccccccccccc}
\hline
\textbf{Model} & \textbf{ar} & \textbf{bn} & \textbf{ca} & \textbf{da} & \textbf{de} & \textbf{en} & \textbf{es} & \textbf{eu} & \textbf{fr} & \textbf{gu} & \textbf{hi}\\
\hline
LLaMA3.1-8B & 46.96 & 36.11 & 51.09 & 48.78 & 51.90 & 57.16 & 54.37 & 32.56 & 53.62 & 48.70 & 43.73 \\
LLaMA3.1-8B-\textsf{AutoCAP} & 41.01 & 42.89 & 50.97 & 50.66 & 53.05 & 49.20 & 50.44 & 27.26 & 52.73 & 43.64 & 38.42 \\
LLaMA3.1-8B-\textsf{QAlign} & 43.47 & 40.46 & 40.67 & 36.62 & 43.40 & 38.31 & 38.53 & 36.05 & 42.44 & 42.68 & 42.30 \\
LLaMA3.1-8B-\texttt{\textbf{AdaMCOT}} & \textbf{54.46} & \textbf{49.68} & \textbf{56.11} & \textbf{50.70} & \textbf{57.49} & \textbf{59.24} & \textbf{57.41} & \textbf{48.45} & \textbf{58.83} & \textbf{49.25} & \textbf{52.78} \\
\hline
Qwen2.5-7B & \textbf{52.78} & 42.25 & 51.87 & 50.32 & 51.27 & 60.59 & 54.88 & 20.80 & 55.53 & 26.81 & 44.11 \\
Qwen2.5-7B-\textsf{AutoCAP} & 44.18 & 33.94 & 45.32 & 43.75 & 44.83 & 53.46 & 47.02 & 16.87 & 50.34 & 22.15 & 35.09 \\
Qwen2.5-7B-\textsf{QAlign} & 45.80 & 27.14 & 46.07 & 47.25 & 52.66 & 59.49 & 52.09 & 15.76 & 52.86 & 17.24 & 33.59 \\
Qwen2.5-7B-\texttt{\textbf{AdaMCOT}} & 47.35 & \textbf{44.81} & \textbf{52.64} & \textbf{51.34} & \textbf{55.33} & \textbf{62.55} & \textbf{56.40} & \textbf{37.21} & \textbf{58.83} & \textbf{31.33} & \textbf{50.06} \\
\hline
\end{tabular}}
\usebox{\myfirsttablebox}

\begin{tabular}{l|ccccccccccc}
\hline
\textbf{Model} & \textbf{hr} & \textbf{hu} & \textbf{hy} & \textbf{id} & \textbf{it} & \textbf{kn} & \textbf{ml} & \textbf{mr} & \textbf{ne} & \textbf{nl} & \textbf{pt}\\
\hline
LLaMA3.1-8B & 49.94 & 47.86 & 32.01 & 46.92 & 54.15 & 48.38 & 44.81 & 47.77 & 43.41 & 51.21 & 57.87 \\
LLaMA3.1-8B-\textsf{AutoCAP} & 48.63 & 43.71 & \textbf{54.79} & 50.51 & 49.55 & \textbf{58.85} & 43.66 & 30.63 & 37.73 & 51.34 & 52.54 \\
LLaMA3.1-8B-\textsf{QAlign} & 38.84 & 41.63 & 37.01 & 42.16 & 39.34 & 41.45 & \textbf{46.54} & 45.81 & 36.82 & 41.53 & 40.61 \\
LLaMA3.1-8B-\texttt{\textbf{AdaMCOT}} & \textbf{57.94} & \textbf{68.04} & 36.35 & \textbf{59.64} & \textbf{57.09} & 49.12 & 45.24 & \textbf{50.39} & \textbf{51.81} & \textbf{57.83} & \textbf{71.45} \\
\hline
Qwen2.5-7B & 45.29 & 46.04 & 18.44 & \textbf{59.13} & 56.07 & 29.06 & 25.07 & 32.20 & 30.62 & \textbf{57.20} & 61.17 \\
Qwen2.5-7B-\textsf{AutoCAP} & 42.15 & 36.72 & 19.95 & 50.13 & 50.86 & 23.57 & 21.18 & 27.25 & 25.06 & 50.34 & 52.61 \\
Qwen2.5-7B-\textsf{QAlign} & 41.88 & 35.15 & 14.31 & 52.31 & 51.85 & 16.96 & 15.71 & 22.38 & 21.73 & 52.74 & 54.06 \\
Qwen2.5-7B-\texttt{\textbf{AdaMCOT}} & \textbf{54.06} & \textbf{69.95} & \textbf{31.65} & 52.44 & \textbf{56.32} & \textbf{36.43} & \textbf{30.26} & \textbf{37.96} & \textbf{42.38} & 55.80 & \textbf{76.93} \\
\hline
\end{tabular}

\begin{tabular*}{\wd\myfirsttablebox}{l|@{\extracolsep{\fill}}cccccccccc}
\hline
\textbf{Model} & \textbf{ro} & \textbf{ru} & \textbf{sk} & \textbf{sr} & \textbf{sv} & \textbf{ta} & \textbf{te} & \textbf{uk} & \textbf{vi} & \textbf{zh}\\
\hline
LLaMA3.1-8B & 53.53 & 44.16 & 48.97 & 48.78 & 53.10 & \textbf{49.39} & 42.84 & 48.57 & 52.48 & 50.13 \\
LLaMA3.1-8B-\textsf{AutoCAP} & 52.76 & 45.30 & 48.20 & 48.28 & 52.84 & 48.32 & 44.54 & 45.06 & 51.97 & 44.04 \\
LLaMA3.1-8B-\textsf{QAlign} & 41.21 & 40.86 & 38.05 & 37.74 & 42.76 & 42.66 & 41.13 & 38.96 & 38.85 & 37.39 \\
LLaMA3.1-8B-\texttt{\textbf{AdaMCOT}} & \textbf{57.25} & \textbf{56.60} & \textbf{60.03} & \textbf{56.61} & \textbf{56.20} & 45.63 & \textbf{45.82} & \textbf{58.05} & \textbf{61.53} & \textbf{59.14} \\
\hline
Qwen2.5-7B & 52.50 & 56.98 & 46.02 & 46.08 & \textbf{56.33} & 23.01 & 22.55 & 52.60 & \textbf{54.90} & \textbf{59.77} \\
Qwen2.5-7B-\textsf{AutoCAP} & 46.92 & 50.81 & 41.67 & 43.25 & 48.71 & 25.90 & 21.43 & 45.93 & 46.66 & 51.72 \\
Qwen2.5-7B-\textsf{QAlign} & 47.37 & 50.13 & 40.49 & 43.57 & 48.32 & 16.55 & 18.87 & 44.29 & 47.39 & 55.39 \\
Qwen2.5-7B-\texttt{\textbf{AdaMCOT}} & \textbf{55.20} & \textbf{60.15} & \textbf{56.04} & \textbf{54.17} & 54.13 & \textbf{31.22} & \textbf{32.77} & \textbf{58.70} & 47.52 & 58.50 \\
\hline
\end{tabular*}

\caption{Multilingual performance comparison on the mTruthfulQA dataset. ISO language codes are used. *-AutoCAP, *-QAlign, and *-AdaMCoT denote the results obtained using the methods from \cite{zhang2024autocap}, \cite{zhu2024question}, and our proposed approach, respectively.}
\label{tab:mtruthfulqa}
\end{table*}

We evaluate \texttt{AdaMCOT} across multiple multilingual benchmarks. For factual accuracy, we use Multilingual TruthfulQA \cite{lai2023okapiinstructiontunedlargelanguage}, covering 31 languages. For open-ended task performance, we use CrossAlpaca-Eval 2.0 \cite{dubois2024length}, featuring parallel questions in English, Chinese, and Indonesian. For reasoning, we assess performance on Cross-MMLU and Cross-LogiQA \cite{wang2024seaevalmultilingualfoundationmodels}, which are adapted from MMLU \cite{hendrycks2021measuringmassivemultitasklanguage} and LogiQA2.0 \cite{10174688}, respectively. Cross-MMLU contains 150 questions and Cross-LogiQA 176, both testing multilingual logical reasoning across English, Chinese, and Indonesian.

\subsubsection{Evaluation Metrics.}
We assess model performance using two distinct evaluation protocols:

\begin{itemize}
    \item \textbf{Open-Ended QA (CrossAlpaca-Eval 2.0):} Responses are rated by GPT-4o (LLM-as-a-judge) on a 0–10 scale for correctness, coherence, and instruction-following, based on strong alignment with human judgments \citep{10.5555/3666122.3668142}.
    \item \textbf{Multiple-Choice QA (TruthfulQA, Cross-MMLU, Cross-LogiQA):} We extract answers using Gemma-2-27B-Instruct \citep{gemmateam2024gemma2improvingopen} and computethe accuracy against the ground truth. For fair comparison, chain-of-thought content is stripped from \texttt{AdaMCOT} outputs before evaluation.
\end{itemize}

We also report cross-lingual consistency ($C$) \cite{wang2024seaevalmultilingualfoundationmodels} for Cross-MMLU and Cross-LogiQA to assess the consistency of response across languages, regardless of correctness of the output.

\subsection{Experimental Results}
\subsubsection{\texttt{AdaMCOT} improves multilingual factual reasoning.}

\begin{table}[t]
\centering
\small 
\setlength{\tabcolsep}{4pt}
\begin{tabular}{l|cccc|cccc}
\hline
 & \multicolumn{4}{c|}{CrossMMLU} & \multicolumn{4}{c}{CrossLogiQA} \\ \cline{2-9}
\textbf{Model} & \textbf{en} & \textbf{zh} & \textbf{id} & \textbf{Cons.} & \textbf{en} & \textbf{zh} & \textbf{id} & \textbf{Cons.} \\ \hline
L-Inst & 82.0 & 68.0 & 67.3 & 66.7 & 61.4 & 54.5 & 46.6 & 51.1 \\ 
L-Ada & \textbf{84.0} & \textbf{74.0} & \textbf{69.3} & \textbf{82.7} & \textbf{64.2} & \textbf{59.7} & \textbf{49.4} & \textbf{71.0} \\ \hline
Q-Inst & 84.0 & 80.7 & 74.7 & 73.3 & 69.3 & \textbf{75.0} & 55.1 & 59.1 \\
Q-Ada & \textbf{84.7} & \textbf{82.7} & \textbf{77.3} & \textbf{91.3}  & \textbf{75.0} & 73.9 & \textbf{62.5} & \textbf{85.8} \\ \hline 
\end{tabular}
\caption{Multilingual performance comparison on CrossMMLU and CrossLogiQA. For each pair of same-family models, the higher value is highlighted. \textit{en}, \textit{zh}, and \textit{id} denote English, Chinese, and Indonesian, respectively. L-Inst, L-Ada, Q-Inst, and Q-Ada refer to LLaMA3.1-8B-Instruction, LLaMA3.1-8B-\texttt{AdaMCOT}, Qwen2.5-7B-Instruction, and Qwen2.5-7B-\texttt{AdaMCOT}, respectively.}
\label{tab:crossmmlulogiqa}
\end{table}

\begin{table}[h!]
\centering
\small
\begin{tabular}{l|c|c|c}
\hline
\textbf{Model} & \textbf{en} & \textbf{zh} & \textbf{id} \\ \hline
LLaMA3.1-8B-Instruction & 8.33 & 7.53 & 7.51 \\
LLaMA3.1-8B-\textbf{\texttt{AdaMCOT}} & \textbf{8.35} & \textbf{8.13} & \textbf{8.13} \\
\hline
Qwen2.5-7B-Instruction & \textbf{8.58} & 8.47 & 7.85 \\
Qwen2.5-7B-\textbf{\texttt{AdaMCOT}} & \textbf{8.58} & \textbf{8.53} & \textbf{8.16} \\ \hline
\end{tabular}
\caption{CrossAlpaca-Eval 2.0: Multilingual Performance Comparison of LLaMA 3.1-8B and Qwen2.5-7B with and without \textbf{\texttt{AdaMCOT}}.}
\label{tab:crossalpaca}
\vspace{-0.3cm}
\end{table}

Table~\ref{tab:mtruthfulqa} shows that \texttt{AdaMCOT} significantly improves performance on TruthfulQA. LLaMA3.1-8B-\texttt{AdaMCOT} boosts accuracy in 31 of 32 languages, with relative gains of 2.1\% in English, 9.0\% in Chinese, and 12.7\% in Indonesian, and over 10\% absolute improvements in low-resource languages like Hungarian, Portuguese, and Bengali. Qwen2.5-7B-\texttt{AdaMCOT} shows similar improvements in 26 languages, notably Basque, Armenian, and Nepali, despite a minor decline in Chinese. These results confirm \texttt{AdaMCOT}’s effectiveness across diverse language typologies.

By contrast, prompt engineering methods (e.g., AutoCAP \cite{zhang2024autocap}) and translation-based approaches \cite{zhu2024question} often fail to enhance, and may even impair, low-resource language performance. The former suffers from ambiguous routing intent and noisy vote aggregation, while the latter introduces semantic drift and depends on scarce parallel data. Detailed reproduction settings and analyses are provided in Appendix A.5.

Reasoning-specific benchmarks (Table~\ref{tab:crossmmlulogiqa}) further validate these findings. LLaMA benefits most in Chinese (+6.0\% CrossMMLU, +5.2\% CrossLogiQA), while Qwen shows the strongest improvement in Indonesian (+2.6\%, +7.4\%). These results highlight \texttt{AdaMCOT}’s ability to bridge gaps introduced by script or training imbalance, particularly in models like LLaMA with Latin-script bias. Even in high-resource languages, both models exhibit consistent gains, indicating that \texttt{AdaMCOT}’s guided reasoning benefits multilingual performance broadly. Additionally, we observe increased cross-lingual answer consistency, suggesting improved semantic alignment.

Table~\ref{tab:crossalpaca} (performance on CrossAlpaca-Eval 2.0) confirms \texttt{AdaMCOT}'s generalizability to open-ended generation. LLaMA improves in Chinese and Indonesian without sacrificing English performance. Qwen likewise maintains its strong baseline in English and Chinese while achieving marked gains in Indonesian. 

Together, these results demonstrate that \texttt{AdaMCOT} transfers reasoning advantages from high- to low-resource languages, reduces hallucination, and improves alignment across both multiple-choice and generative tasks.

\begin{table}[t]
\centering
\small
\begin{tabular}{p{5.2cm}|p{0.5cm}<{\centering}p{0.5cm}<{\centering}p{0.5cm}<{\centering}}
\hline
 & \multicolumn{3}{c}{CrossAlpaca-Eval 2.0} \\ \cline{2-4}
\textbf{Model} & \textbf{en} & \textbf{zh} & \textbf{id} \\ 
\hline
LLaMA3.1-8B (Baseline) & 8.33 & 7.53 & 7.51 \\
LLaMA3.1-8B-\texttt{AdaMCOT} (Direct) & 8.35 & 7.38 & 7.59 \\
LLaMA3.1-8B-\texttt{AdaMCOT} (English) & 8.35 & 7.71 & 7.86 \\
LLaMA3.1-8B-\texttt{AdaMCOT} (Chinese) & 6.99 & 7.38 & 7.23 \\
LLaMA3.1-8B-\texttt{AdaMCOT} (Indonesian) & 7.11 & 7.21 & 7.59 \\
LLaMA3.1-8B-\texttt{AdaMCOT} (w/o Filter) & 8.23 & 8.07 & 8.11 \\
LLaMA3.1-8B-\texttt{AdaMCOT} & \textbf{8.35} & \textbf{8.13} & \textbf{8.13} \\
\hline
\end{tabular}
\caption{Ablation study results on CrossAlpaca-Eval 2.0. We report the mean GPT-4o scores for English (\textbf{en}), Chinese (\textbf{zh}), and Indonesian (\textbf{id}). 
``Direct'' indicates no intermediate reasoning, while ``English,'' ``Chinese,'' or ``Indonesian'' indicates intermediate reasoning in that language. ``w/o Filter'' applies adaptive routing but omits score-based filtering.}
\label{tab:crossalpaca_ablation}
\vspace{-0.3cm}
\end{table}

\subsubsection{Adaptive Language Routing for Enhanced \texttt{AdaMCOT} Performance.}
Adaptive Language Routing (ALR), a core mechanism in the \texttt{AdaMCOT} framework, enhances multilingual factual reasoning by dynamically selecting the optimal intermediate language for each instruction, guided by input characteristics and performance feedback.

To systematically investigate the influence of various language routing strategies, we performed an ablation study on the CrossAlpaca-Eval 2.0 benchmark. This dataset enables a comprehensive exploration of routing strategies across diverse instruction types. Our ablation compares five variants of our method applied to LLaMA 3.1-8B, alongside the official LLaMA 3.1-8B baseline and the original \texttt{AdaMCOT}. The mean GPT-4o scores for all three primary languages are summarized in Table \ref{tab:crossalpaca_ablation}.

Our analysis first investigates the effect of reasoning in a single, fixed language and then evaluates the effectiveness of a score-based filtering mechanism. We observe the following:
(1)The model's extensive English pre-training makes English-only reasoning consistently outperform other languages, establishing English as the dominant knowledge resource in LLaMA.
(2) The score-based filtering mechanism improves ALR's performance, demonstrating the importance of retaining only high-quality training examples.
(3) ALR significantly surpasses a fixed-English strategy with relative performance gains of 5.2\% and 9.3\%, underscoring the need for dynamic reasoning to maximize performance.

\subsubsection{Reasoning Pathway Distribution under the \texttt{AdaMCOT} Framework.}

We analyzed the impact of different reasoning paths chosen by \texttt{AdaMCOT} across datasets. Figure~\ref{fig:changes} shows the routing distribution for LLaMA3.1-8B-\texttt{AdaMCOT} on the mTruth dataset. A consistent pattern emerges: the model strongly favors high-resource languages, especially English, as intermediate reasoning paths, followed by direct generation. This trend holds for both beneficial (incorrect-to-correct) and detrimental (correct-to-incorrect) outcome shifts. Despite being a high-resource language, Chinese is selected less frequently due to its underrepresentation in LLaMA’s pretraining corpus. Indonesian, while low-resource, occasionally contributes positively, indicating that effective prompting can surface useful knowledge even from underrepresented languages.

In general, adaptive routing yields more correct outputs, with beneficial pathways outnumbering harmful ones. Importantly, \texttt{AdaMCOT} sometimes selects languages outside the instruction-tuning set, such as Marathi, Serbian, and Nepali, and still achieves positive results. This highlights its generalizability in identifying effective reasoning paths beyond the predefined auxiliary languages. Additional distributions for other tasks are included in Appendix A.3.

\subsubsection{Case Studies on \texttt{AdaMCOT}.}
 
We conduct a more detailed case study to illustrate how \texttt{AdaMCOT} dynamically selects between direct generation and intermediate reasoning, optimizing multilingual performance across diverse tasks without compromising answer quality. In linguistically dependent tasks like composing a Chinese poem, the model strategically generates content directly in Chinese, leveraging the language's inherent semantic richness to preserve poetic fluency and avoid the potential information loss associated with translation or intermediate reasoning processes. Likewise, when prompted with an English word riddle that simply asks for a rhyming word, \texttt{AdaMCOT} again employs direct generation and provides the correct answer on par with the baseline. Notably, in both cases, the \texttt{AdaMCOT} fine-tuned model shows no degradation in answer quality.

For questions where intermediate reasoning can help, \texttt{AdaMCOT} leverages high-resource languages (e.g., English) to boost accuracy. An example is the Chinese probability question, where the base LLaMA3.1-8B model incorrectly predicts the chance of getting at least one head in two coin tosses, but \texttt{AdaMCOT}’s chain-of-thought in English yields the correct $3/4$ answer in Chinese. Similarly, when asked in Indonesian about Singapore’s longest expressway, the baseline mistakenly identifies the KPE, while \texttt{AdaMCOT} correctly names the PIE by tapping into its richer English-based knowledge. These examples underscore the  \texttt{AdaMCOT}'s adaptive reasoning approach, highlighting its capacity to dynamically select optimal linguistic pathways and significantly improve multilingual reasoning across diverse task domains while maintaining performance in high-resource language setting. The effectiveness of using Indonesian as a thinking language is evident in completing Pantun, where reasoning in Indonesian yields responses that better follow its traditional structure. This shows that for culturally specific tasks, low-resource languages can sometimes enable superior reasoning. We provide more specific examples, error and limitation analyses in Appendix A.7.

\subsection{Interpretability Study of \texttt{AdaMCOT}}

Prior research suggests that the reasoning processes of multilingual LLMs primarily occur within a shared, language-agnostic latent space. This shared space allows models to perform reasoning tasks across different languages, depending on the quality of their multilingual alignment. These reasoning dynamics are especially prominent in the middle and upper layers of the model, closer to the output layer \cite{zhao2024large,schut2025multilingualllmsthinkenglish,wang2025lostmultilingualitydissectingcrosslingual,fierro2025multilinguallanguagemodelsremember}. Furthermore, studies on knowledge neurons and cross-lingual activation patterns indicate that stronger multilingual alignment improves factual transfer and consistency \cite{wang2024sharing,tang-etal-2024-language,cao2024one}. To better understand how \texttt{AdaMCOT} enhances multilingual reasoning, we conduct an interpretability analysis using the Logit Lens \cite{nostalgebraist2020logitlens} and UMAP \cite{mcinnes2020umapuniformmanifoldapproximation}. 

\subsubsection{Logit Lens Analysis.}
We use the Logit Lens to examine the model’s layer-wise hidden states by projecting them onto the output vocabulary, revealing its predictive focus at each decoding step. In Figures~\ref{fig:llama-adop-hot-15} and~\ref{fig:llama-adop-hot-88-103}, each row represents a model layer and each column a decoding step; only the token with the highest logit is visualized to trace dominant predictions. To illustrate the impact of cross-lingual reasoning, we analyze responses to: \textit{“Which countries have longer working hours than Mexico?”}. As shown in Figure~\ref{fig:answer}, the baseline LLaMA3.1-8B-Instruction model hallucinates facts when reasoning directly in Chinese, likely due to insufficient Chinese training data.

By contrast, the \texttt{AdaMCOT}-enhanced model first reasons in English (Figure~\ref{fig:llama-adop-hot-15}). Although early predictions are noisy, it ultimately converges on the correct answer. It then generates the final response in Chinese, grounded in the English chain-of-thought (Figure~\ref{fig:llama-adop-hot-88-103}). This shift leads to more confident and coherent token selection, demonstrating that routing reasoning through high-resource languages enhances factual accuracy and reduces uncertainty.

\subsubsection{UMAP Embedding Analysis.}
To further investigate cross-lingual semantic alignment, we apply UMAP to visualize the model's language-specific embedding distributions before and after applying \texttt{AdaMCOT}. We observe that AdaMCoT brings language-specific embeddings closer, particularly around the English centroid. For instance, in the LLaMA3.1-8B model, the average distance from non-English clusters to the English centroid decreases from 9.87 to 9.39 after applying \texttt{AdaMCOT}. Importantly, this alignment occurs without significant distortion of the original embedding space. As all training data are generated by the baseline model, this shift reflects \texttt{AdaMCOT}’s ability to promote better multilingual coherence without catastrophic forgetting. Similar improvements are observed in Qwen2.5 models, confirming the generalizability of the effect. Detailed UMAP and Logit Lens plots are available in Appendix A.4.

\section{Conclusion}
We presented \texttt{AdaMCOT}, a novel framework for enhancing cross-lingual factual reasoning in LLMs via adaptive chain-of-thought prompting. By dynamically routing reasoning through strategically chosen ``thinking languages,'' \texttt{AdaMCOT} mitigates performance disparities across languages, especially improving outcomes in low-resource settings, while preserving or improving accuracy in high-resource ones. Our method introduces a reward-based training procedure that selects optimal reasoning pathways using a strong LLM-based evaluator, enabling the model to learn when and how to reason cross-lingually. Comprehensive evaluations across four multilingual benchmarks demonstrate consistent gains in both factual accuracy and cross-lingual consistency. Finally, our interpretability analysis using Logit Lens and UMAP reveals that \texttt{AdaMCOT} promotes better semantic alignment across languages and reduces reasoning hallucinations. These findings suggest that adaptive language routing is a promising direction for improving multilingual LLM performance without requiring additional pretraining or translation-heavy pipelines.

\bibliography{aaai2026}

\newpage
\appendix
\onecolumn

\begin{CJK}{UTF8}{gbsn}
\section{Appendix}
\subsection{Related Work}
Effective multilingual alignment requires careful design of both pretraining and instruction tuning strategies \cite{gao-etal-2024-multilingual}. Recent approaches include multilingual contrastive learning with cross-lingual instruction tuning \cite{li-etal-2024-improving-context} and word-level alignment before code-switched pretraining \cite{li2024prealign}. Translation parallel pairs have proven particularly effective \cite{zhang2023bayling,zhu2024question,mu2024large,lin2024recipe}. Notable advances include two-stage training combining extensive monolingual data with high-quality translation datasets \cite{xuparadigm}, integration of multilingual translation and task instructions \cite{zhu2023extrapolating,lin2024crossin}, and self-distillation methods transferring capabilities from high to low-resource languages \cite{zhang-etal-2024-enhancing-multilingual}. While cross-lingual experts train independently on multilingual corpus subsets \cite{blevins-etal-2024-breaking}, native language prompting remains crucial for culture-specific tasks requiring nuanced understanding \cite{liu2024translation}. 
For cross-lingual reasoning, multilingual instruction tuning with English as the default thinking language improves reasoning consistency on math problems \cite{lai-etal-2024-mcot}. Moreover, layer transplantation between language and math experts, fine-tuned from a common pretrained model, enhances mathematical performance in target languages \cite{bandarkar2024layer}.

\subsection{Exploring Reward Model Selection, Consistency with Human Preferences, and Associated Risks}

\subsubsection{Reward Model Selection}
We conducted a series of human evaluations on GPT-4o’s multilingual capabilities to ensure its suitability as a reward model for our task. The evaluation consisted of the following components: 
\begin{enumerate}
    \item The accuracy of machine translation for low-resource languages;
    \item The ability to paraphrase sentences;
    \item The model’s reasoning ability.
\end{enumerate}

The evaluation of translation accuracy aimed to assess the model’s cross-lingual mapping capability, ensuring that GPT-4o could accurately produce answers in one language based on intermediate reasoning in another. We selected English, Malay, Tamil, Indonesian, Vietnamese, Chinese, and French as the evaluation languages. For each language, 50 sentences were sampled from the Flores-200 dataset~\cite{nllb2022}, and one language was randomly chosen as the target for GPT-4o to translate into. The outputs were manually evaluated by language experts. Results showed that while GPT-4o occasionally produced incomplete or unnatural translations in low-resource languages, over 92\% of the translations are correct.

The sentence paraphrasing task was designed to test the model’s comprehension across different languages. Using the same set of sentences as in the translation task, we asked GPT-4o to rewrite them. Human evaluation indicated that GPT-4o performed excellently in this task, demonstrating strong understanding of multilingual content.

The model’s reasoning ability is discussed in detail in the technical report~\cite{openai2024gpt4ocard} and will not be elaborated here. Based on these evaluations, we conclude that GPT-4o is well-suited to serve as the reward model in our work.

\subsubsection{Consistency with Human Preferences and Associated Risks}
To mitigate the risk of GPT-4o introducing significant discrepancies from human preferences when used as a reward model, we conducted a human consistency study evaluating GPT-4o’s performance on open-ended QA tasks. Specifically, we randomly sampled 50 evaluations made by GPT-4o on model responses to questions from the Alpaca dataset. For each evaluation, three language experts independently voted on whether they agreed with GPT-4o’s assessment of the model output. The results showed a high consistency rate of 86\% between GPT-4o and human preferences. Although some disagreements remain, we believe the overall consistency meets the standard required for use as a reward model. 
Meanwhile, GPT-4o has strong restrictions in terms of ethical and safety concerns, and our task is primarily focused on fact-based or fixed-premise question answering, which generally does not involve such risks. Although using GPT-4o may introduce stylistic preferences in its responses, the impact on our task outcomes is limited. We leave deeper analysis of this aspect to future work.

\subsection{Distribution of Reasoning Pathway Selections}

\begin{figure}[H]
  \centering
  \includegraphics[width=1\textwidth]{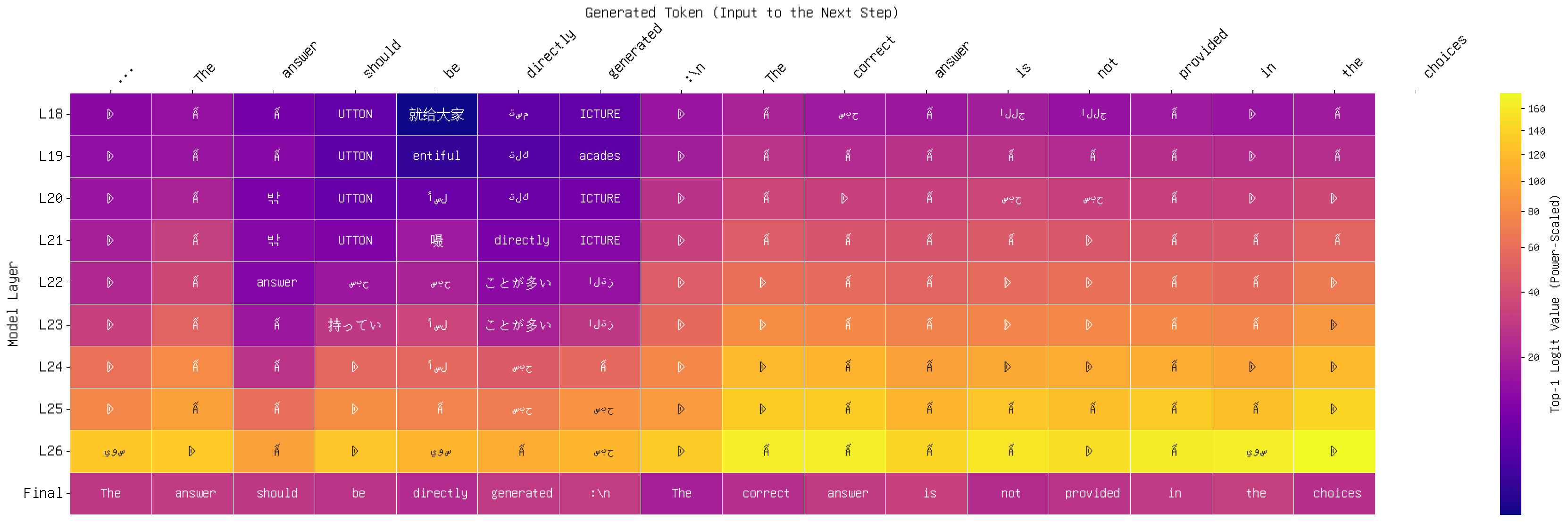}
  \caption{Logit lens maps for Qwen2.5-7B-\textit{AdaMCOT}, when prompted with question \textit{Which animals can add up numbers just as well as humans? A: No animals can add up numbers
just as well as humans ... K: Parrots can add up numbers just as well as humans.}} 
  \label{fig:qwen_adop_1_15}
\end{figure}

As shown in Figure \ref{fig:changes}, after applying AdaMCOT to Qwen2.5, the model predominantly selects direct answering as its language routing strategy, followed closely by reasoning in English and Chinese. This behavior stems from Qwen2.5’s highly aligned and tightly intertwined semantic space, which is clearly illustrated in the intermediate state visualizations in Figure \ref{fig:qwen_adop_1_15}. Unlike LLaMA-3.1, which tends to perform intermediate reasoning in a specific language as showing in Fig \ref{fig:llama_1_15}, Qwen2.5 operates in a more language-agnostic shared reasoning space. This suggests that during direct answering, the model may implicitly perform cross-lingual mapping without making it linguistically explicit. Nevertheless, our method significantly improves the accuracy of direct answers, likely due to its facilitation of cross-lingual knowledge transfer, which helps enhance the model's implicit cross-lingual reasoning capabilities.

\begin{figure*}[h]
  \centering
  \includegraphics[trim=0 100 0 0, clip, width=0.85\textwidth]{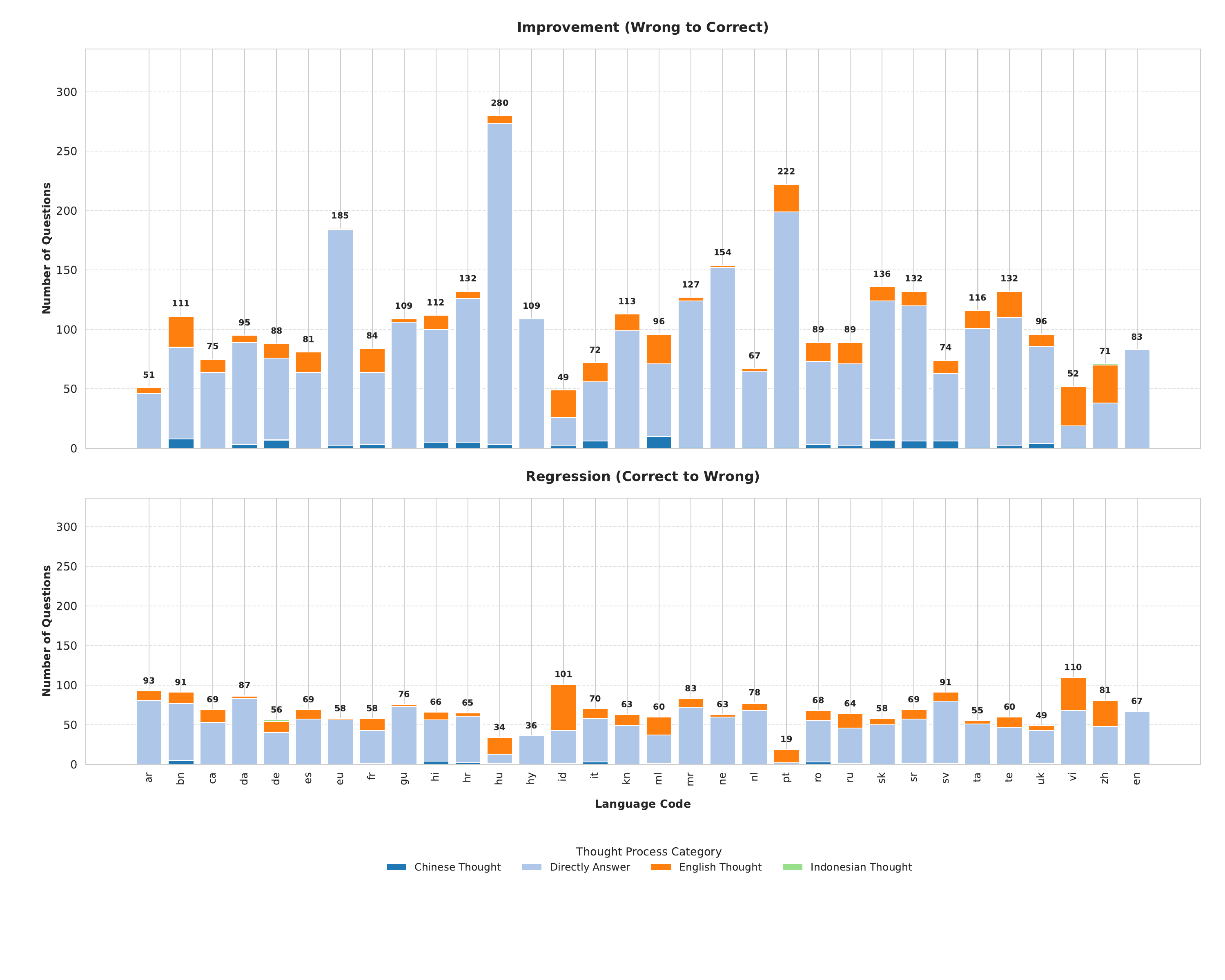}
  \caption{Distribution of Reasoning Pathway Selections on the mTruth Dataset: Qwen2.5-7B-AdaMCOT vs. Base Model.} 
  \label{fig:qwen-changes}
\end{figure*}

We further analyzed the distribution of reasoning pathway selections on the CrossMMLU and CrossLogiQA datasets before and after applying \textit{AdaMCOT}. As illustrated in Figure \ref{fig:qwen-changes}, the observed patterns exhibit slight deviations from those on the mTruth dataset. Specifically, after integrating AdaMCOT, LLaMA demonstrates a clear preference for high-resource languages—particularly English—as the primary reasoning language in tasks involving logical inference and comprehension. This trend can be attributed to the superior logical reasoning capabilities commonly associated with English. Unlike factual question answering tasks, where Chinese previously served as a secondary reasoning path, direct answering has emerged as the preferred alternative to English. This observation echoes prior findings by ~\citet{zhao2024large}, suggesting that reasoning abilities in large language models may be inherently shared across languages.

In contrast, Qwen exhibits a consistent reasoning pathway distribution across all datasets, reinforcing its internally aligned multilingual representation. This is further evidenced in Figure \ref{fig:qwen_adop_1_15}, where Qwen's intermediate representations appear to form a language-agnostic latent space—most reasoning steps are not explicitly associated with any particular language.

\begin{figure}[H]
  \centering
  \includegraphics[width=0.85\textwidth]{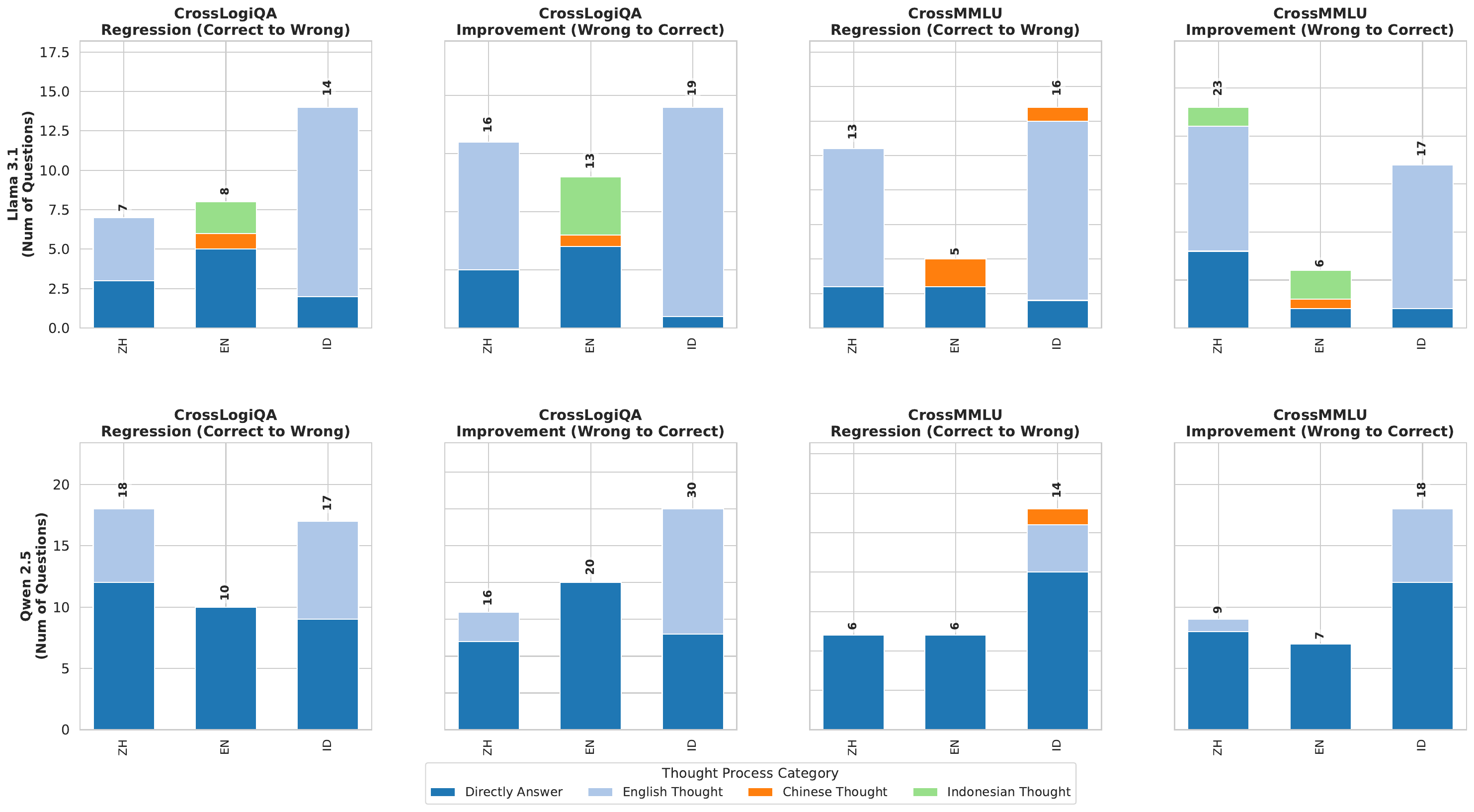}
  \caption{Distribution of Reasoning Pathway Selections on the CrossMMLU \& CrossLogiQA Dataset: Qwen2.5-7B-AdaMCOT and Llama-3.1-8B-AdaMCOT.} 
  \label{fig:changes_cross}
\end{figure}

\subsection{Interpretability Study of AdaMCOT}

\begin{figure*}[htbp]
  \centering
  \includegraphics[trim=0 20 0 5, clip, width=1\textwidth]{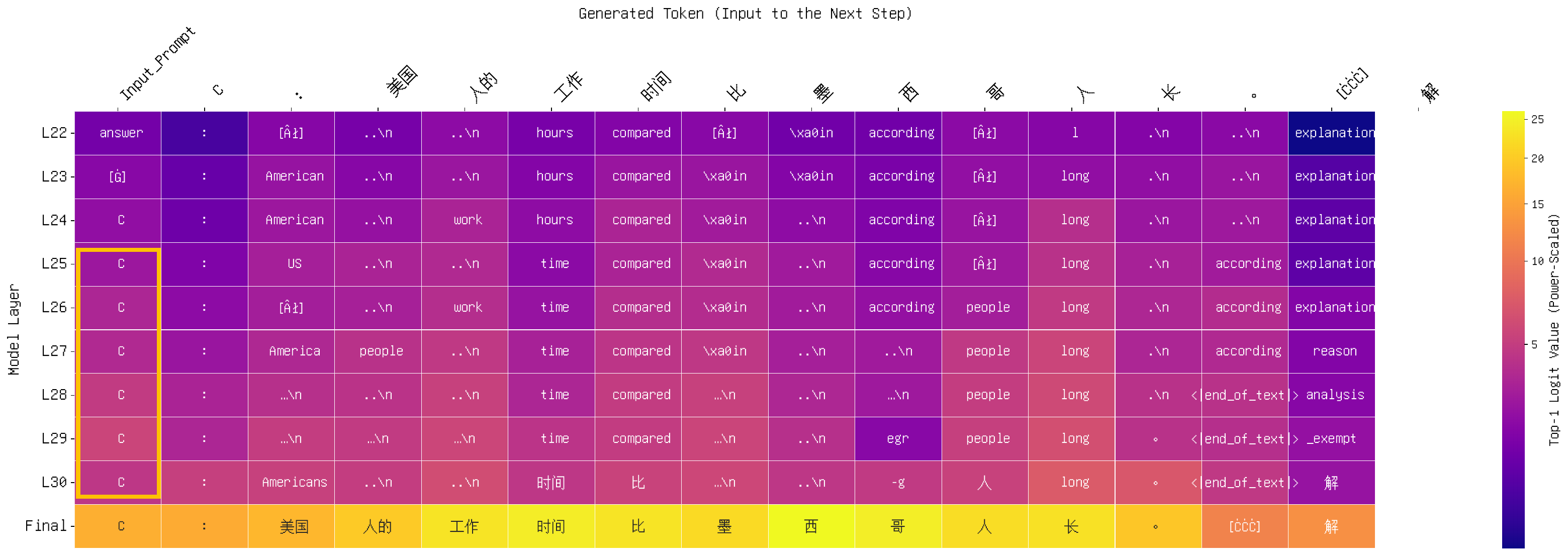}
  \caption{Logit lens maps for Llama-3.1-8B-Instruction, when prompted with question \textit{``哪些国家的工作时间比墨西哥人更长？A: 墨西哥人的工作时间很长，但柬埔寨可能更长。... F: 墨西哥人的工作时间比中国人短。(Which countries have longer working hours than Mexico? A.Mexicans work long hours, but Cambodians might work even longer ... F: Mexicans work fewer hours than the Chinese.)"}} 
  \label{fig:llama_1_15}
\end{figure*}

Figure \ref{fig:llama_1_15} presents the intermediate layer representations of the model visualized using the Logit Lens, based on the prompt: “哪些国家的工作时间比墨西哥人更长？(Which countries have longer working hours than Mexico?)”. As illustrated in the visualization, the model consistently converges on an incorrect answer when attempting to directly respond to the question. This behavior may be attributed to the fact that LLaMA’s training data is predominantly composed of Latin-based languages, with relatively limited coverage of Chinese, resulting in potential knowledge gaps in Chinese-language contexts. However, such hallucinations were effectively mitigated following the application of AdaMCOT, which utilizes English as the intermediate reasoning language due to its extensive linguistic knowledge base.

\begin{figure}[H]
  \centering
  \includegraphics[width=1\textwidth]{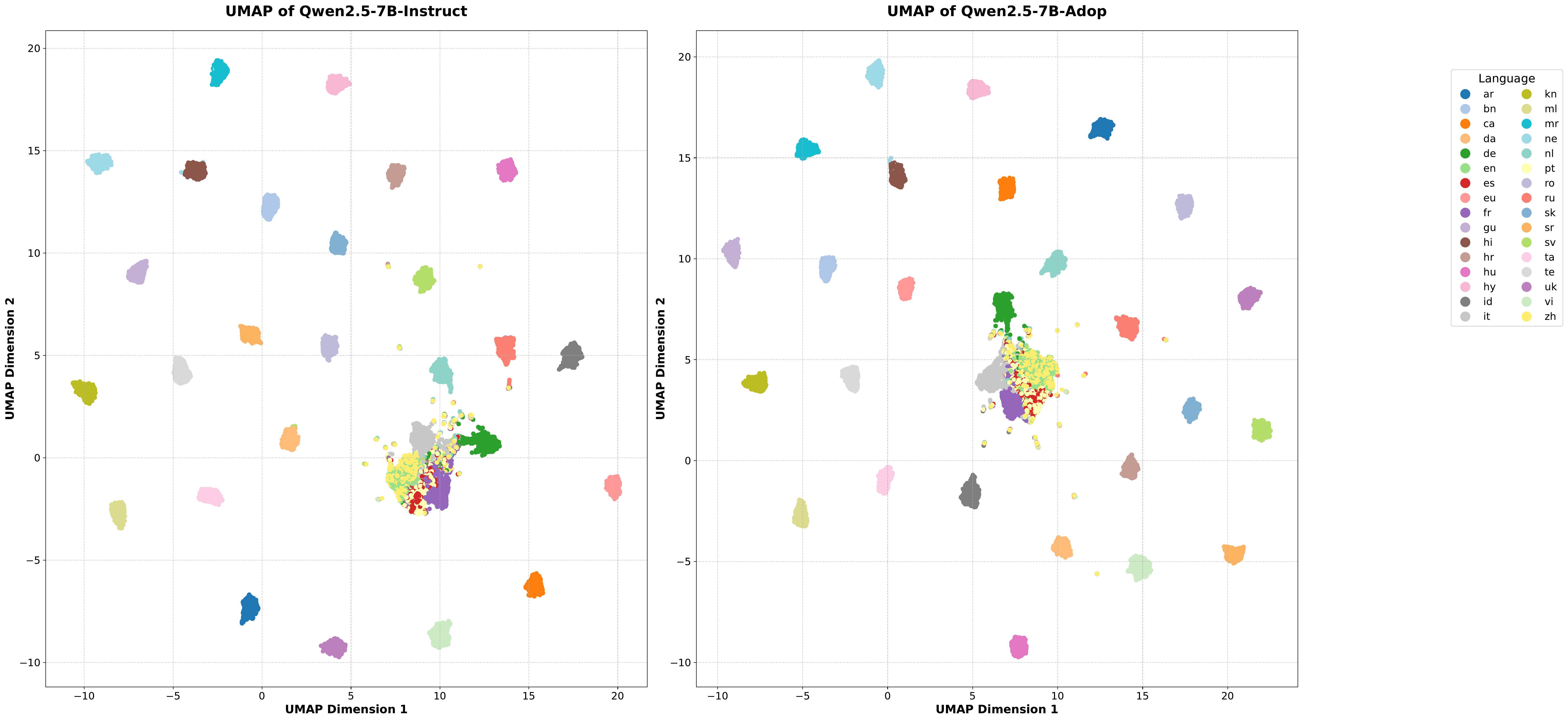}
  \caption{UMAP Visualization of Semantic Spaces of
  Qwen-2.5-7B-Instruction and Qwen-2.5-7B-AdaMCOT} 
  \label{fig:qwen-compare}
\end{figure}

A similar reasoning calibration can also be observed in Qwen. As demonstrated in Table \ref{tab:Qwen-Answer}, when answering the question “Which animals can add up numbers just as well as humans? A: No animals can add up numbers just as well as humans. B: Monkeys can add up numbers just as well as humans. C: Chimpanzees can add up numbers just as well as humans. D: Apes can add up numbers just as well as humans. E: Dogs can add up numbers just as well as humans. F: Birds. G: Dogs. H: Apes. I: Chimps. J: Birds can add up numbers just as well as humans. K: Parrots can add up numbers just as well as humans.” Qwen-AdaMCOT, despite opting for direct answering, exhibits more coherent reasoning than the base model and ultimately arrives at the correct answer. These findings indicate that our method not only optimizes the selection of intermediate reasoning languages but also enhances the model's overall reasoning capability. Furthermore, as illustrated in Figure \ref{fig:qwen-compare}, the semantic distance between English and non-English representations within Qwen becomes smaller after applying AdaMCOT, decreasing from 11.19 to 10.05. This closer alignment facilitates cross-lingual knowledge transfer and positively contributes to the model’s fact-based reasoning performance.


\begin{table}[H] 
\small
\centering 
\begin{tabular}{@{} l | p{0.8\textwidth} @{}}
\toprule
\textbf{Model} & \textbf{Answer} \\
\midrule

Qwen-Adop &
The answer should be directly generated:
The correct answer is not provided in the choices. However, based on scientific research, the correct answer would be:

A: No animals can add up numbers just as well as humans.

While some animals, such as chimpanzees, apes, and some birds, have been shown to have a basic understanding of numbers and can perform simple arithmetic tasks, they do not have the same level of numerical ability as humans. Humans have a unique ability to understand and manipulate numbers in a way that is not found in other animals.
 \\

\midrule

Qwen-IT &
None of the given choices accurately reflect the current scientific understanding of animal numerical abilities. While some animals, particularly primates, have shown impressive skills in numerical tasks, they do not perform these tasks at the same level as humans. Here's a more accurate assessment based on the current scientific literature:

- \textbf{Chimpanzees (I)} and \textbf{Apes (H)}: Chimpanzees and other apes have demonstrated the ability to understand and manipulate small numbers, and they can perform simple addition tasks, but their abilities are limited compared to humans. They can recognize numerical quantities and perform basic arithmetic, but they do not "add up numbers just as well as humans."

- \textbf{Monkeys (B)}: Some species of monkeys, like capuchin monkeys, have also shown some ability to understand numerical concepts and perform simple addition tasks, but again, their abilities are not comparable to those of humans.

- \textbf{Dogs (G) and (Dogs) (G)}: Dogs have shown some understanding of numerical concepts, such as recognizing the difference between a small number of objects, but they do not perform addition tasks as complex as those performed by humans.
- \textbf{Birds (F) and (Birds) (F)}: Some bird species, like African grey parrots, have demonstrated the ability to understand and manipulate small numbers and even perform simple addition tasks, but their abilities are still limited compared to humans.

- \textbf{Parrots (J) and (Parrots) (J)}: Similar to birds, some parrots have shown the ability to understand and manipulate small numbers, but they do not perform addition tasks as complex as those performed by humans.

Therefore, none of the given choices accurately describe animals that can add up numbers just as well as humans. The closest would be the ability of some primates and birds to understand and manipulate small numbers, but this is far from performing addition tasks as complex as those performed by humans.
\\

\bottomrule
\end{tabular}
\caption{Answers of Qwen-2.5-7B-Instruction and Qwen-2.5-7B-\textsf{AdaMCoT}, when prompted with question \textit{Which animals can add up numbers just as well as humans? A: No animals can add up numbers just as well as humans ... K: Parrots can add up numbers just as well as humans}. Qwen-Adop and Qwen-IT refer to Qwen2.5-7B-Instruction and Qwen2.5-7B-\textsf{AdaMCOT}.}
\label{tab:Qwen-Answer} 
\end{table}

\begin{figure}[H]
  \centering
  \includegraphics[width=1\textwidth]{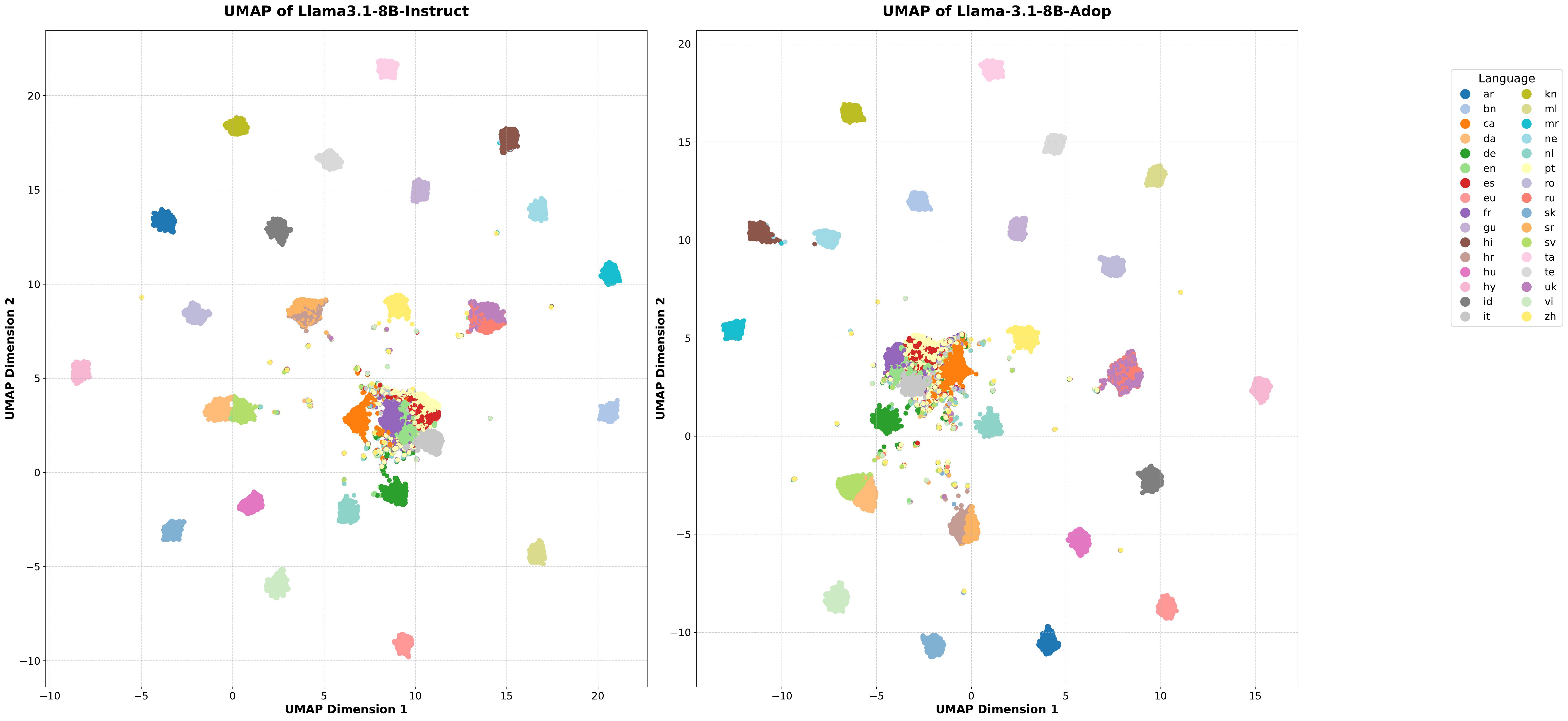}
  \caption{UMAP Visualization of Semantic Spaces of
  Llama-3.1-8B-Instruction and Llama-3.1-8B-AdaMCOT} 
  \label{fig:llama-compare}
\end{figure}

\subsection{Baseline reproduction details and result analysis}

We reproduced two baselines in this work, and the following are the reproduction details.

\subsubsection{AutoCAP~\cite{zhang2024autocap}} We followed the method described in the paper, setting the number of automatically routed languages to 6. The prompts used at each stage are shown in the Table \ref{table:prompt}. All prompt templates strictly follow the content presented in the original paper. For the model-generated responses, we use GPT-4o to extract the final option from the output statement, ensuring the accuracy of the extracted result.

\begin{table*}[h!]
\centering
\scriptsize
\renewcommand{\arraystretch}{1.3}
\begin{tabularx}{\textwidth}{l | l| X}
\toprule
\textbf{Stage} & \textbf{Model} & \textbf{Prompt} \\
\midrule

\multirow{2}{*}{Automatic Language Selection Prompting} 
& LLaMA3.1-8B-AutoCAP & 
\texttt{[Language Information] The language information of training data for LLaMA is as follows: English: ~65\%. Group 1 (~10\%): German, French, Spanish, Italian, Portuguese. Group 2 (~5\%): Dutch, Polish, Swedish. Group 3 (~10\%): Chinese, Japanese, Hindi, Vietnamese, Turkish. Task: Select the top 6 languages suitable for reasoning on the following query in \{source\_lang\}. Query: \{query\}. Output format: Comma-separated list of languages. For example: English, Chinese, Japanese. Don't output any other additional information.} \\
\cline{2-3}

& Qwen2.5-7B-AutoCAP & 
\texttt{[Language Information] The language information of training data for Qwen is as follows: Chinese: ~60.5\%, English: ~23.8\%, Other High-Resource Languages: ~8.1\% (German, French, Spanish, Italian, Portuguese, Dutch, Russian, Japanese, Korean). Task: Select the top 6 languages suitable for reasoning on the following query in \{source\_lang\}. Query: \{query\}. Output format: Comma-separated list of languages. Don't output any other additional information.} \\

\midrule

\multirow{2}{*}{Automatic Weight Allocation Prompting} 
& LLaMA3.1-8B-AutoCAP & 
\texttt{You have selected the following languages for reasoning: \{lang\_str\}. Task: Assign a confidence score (0.0 to 1.0) for each language based on their relevance to the query "\{query\}". Format: One line per language in the form "Language: Score". For example: English: 0.8, Chinese: 0.2. Don't output any other additional information.} \\
\cline{2-3}

& Qwen2.5-7B-AutoCAP & 
\texttt{You have selected the following languages for reasoning: \{lang\_str\}. Task: Assign a confidence score (0.0 to 1.0) for each language based on their relevance to the query "\{query\}". Format: One line per language in the form "Language: Score". For example: English: 0.8, Chinese: 0.2. Don't output any other additional information.} \\

\midrule

\multirow{2}{*}{Reasoning} 
& LLaMA3.1-8B-AutoCAP & 
\texttt{Please reason step-by-step to solve the following question in \{lang\}. Query: \{query\}. Answer:} \\
\cline{2-3}

& Qwen2.5-7B-AutoCAP & 
\texttt{Please reason step-by-step to solve the following question in \{lang\}. Query: \{query\}. Answer:} \\

\bottomrule
\end{tabularx}
\caption{Prompt templates used in AutoCAP for language selection, weight allocation, and reasoning across models.}
\label{table:prompt}
\end{table*}

\subsubsection{QAlign~\cite{zhu2024question}} 
For fine-tuning the model on translation tasks, we followed the original setup by using the parallel instructions extracted by the authors from the \textit{GSM8KInstruct\_Parallel}~\cite{chen2023breaking} dataset. We performed LoRA fine-tuning ~\cite{hu2022lora} for 1 epoch with a learning rate of 5e-5 and a LoRA rank of 32. The prompt template for the translation task also follows the original format: \textit{[Source Language] {Source Sentence} [Target Language]}, and the expected output is the corresponding target sentence.

For the instruction fine-tuning stage, since the original work used the English mathematical reasoning dataset \textit{METAMATHQA} for the instruction fine-tuning, which differs from our more diverse tasks, we used the Alpaca~\cite{alpaca} and GPT-4 Instruction datasets~\cite{peng2023instruction} for fair comparison. Instruction fine-tuning was also conducted for 1 epoch with a learning rate of 5e-5 and a LoRA rank of 32. All fine-tuning experiments were conducted on 8 NVIDIA H100 80GB GPUs.

\subsubsection{Analysis} 

Through our analysis of the AutoCAP results, we observed that prompt engineering-based approaches exhibit a degree of randomness in selecting intermediate reasoning languages. This randomness often introduces noise, as the model's reasoning ability in many low-resource languages may be limited. As a result, the aggregated answer based on weighted voting can amplify incorrect reasoning paths, increasing the likelihood of selecting the wrong answer over the correct one. Moreover, the model consistently assigns higher weights to intermediate languages that match the prompt language, indicating that without fine-tuning, prompt engineering alone struggles to overcome the model’s inherent language bias during inference. Additionally, this approach requires multiple forward passes, significantly reducing inference efficiency.

The translation-based approach presents several challenges. First, parallel corpora between low-resource languages and English are scarce and difficult to acquire. Second, noise introduced by translation errors can negatively impact model performance. Furthermore, our observations indicate that for models with strong multilingual capabilities—such as Qwen2.5—the benefits of translation-based methods are marginal, as these models already exhibit a substantial degree of cross-lingual alignment.

\subsection{Training details}

Our training was conducted on 8 NVIDIA H100 (80GB) GPUs. For the instruction fine-tuning, the training time for each model was approximately 7 to 7.5 hours. All training was performed using bf16 precision with Low-Rank Adaptation (LoRA) \cite{hu2022lora}. The specific training parameters are provided in Table \ref{table:instruction-finetuning parameters}.

\begin{table}[ht!]
\centering
\small
\renewcommand{\arraystretch}{1.1}
\begin{tabular}{p{4.0cm}p{2.0cm}}
\toprule
\textbf{Parameter} & \textbf{Value} \\
\midrule
Training Batch Size & 256 \\
Max Training Epochs & 1 \\
Warm Up Steps & 500 \\
Lora Rank & 32 \\
Lora Alpha & 64 \\
Maximum tokens & 3000 \\
Learning rate & 5e-5 \\
\bottomrule
\end{tabular}
\caption{Training hyperparameters for the instruction fine-tuning}
\label{table:instruction-finetuning parameters}
\vspace{-0.3cm}
\end{table}

\subsection{Case Study}

\subsubsection{Success Case Study}
As shown in Table \ref{tab:adaptive_reasoning_case_studies}, in linguistically dependent tasks like composing a Chinese poem, the model strategically generates content directly in Chinese, leveraging the language's inherent semantic richness to preserve poetic fluency and avoid the potential information loss associated with translation or intermediate reasoning processes. A comparison between the outputs of LLaMA3.1-8B-AdaMCOT and LLaMA3.1-8B-Instruction reveals that the application of AdaMCOT results in more rhythmically poetic compositions, whereas the outputs from LLaMA3.1-8B-Instruction tend to be more prosaic and straightforward. Likewise, when prompted with an English word riddle that simply asks for a rhyming word, \textsf{AdaMCoT} again employs direct generation and provides the correct answer on par with the baseline. Notably, in both cases, the \textsf{AdaMCoT} fine-tuned model shows no degradation in answer quality.

For questions where intermediate reasoning can help, \textsf{AdaMCoT} leverages high-resource languages (e.g., English) to boost accuracy. An example is the Chinese probability question, where the base LLaMA3.1-8B model incorrectly predicts the chance of getting at least one head in two coin tosses, but \textsf{AdaMCoT}’s chain-of-thought in English yields the correct $3/4$ answer in Chinese. Similarly, when asked in Indonesian about Singapore’s longest expressway, the baseline mistakenly identifies the KPE, while \textsf{AdaMCoT} correctly names the PIE by tapping into its richer English-based knowledge. These examples underscore the  \textsf{AdaMCoT}'s adaptive reasoning approach, highlighting its capacity to dynamically select optimal linguistic pathways and significantly improve multilingual reasoning across diverse task domains while maintaining performance in high-resource language setting.

\begin{table}[H]
\centering
\scriptsize
\begin{tabular}{P{4cm}|P{2cm}|P{4cm}|P{4cm}}
\hline
\textbf{Input Query} & \textbf{Pathway} & \textbf{AdaMCOT} & \textbf{Base Model} \\
\hline
创作一首秋天的诗歌 
\textcolor{gray}{(Compose a poem about autumn)} 
& Direct Generation
& 秋天的风，带来凉爽的气息; 落叶的黄金，铺满了大地的面貌\textcolor{gray}{(The autumn wind brings a breath of coolness;
Golden fallen leaves blanket the face of the earth.)}
& 秋天的叶子飘落下，金黄的色彩在空中舞。风轻轻地吹过，带来凉爽，秋天的美景，永远留在心头。\textcolor{gray}{(Autumn leaves gently fall, golden hues dancing through the air. A soft breeze blows, bringing a cool touch, The beauty of autumn; Forever lingers in the heart.)}
\\
\hline
I’m a word that rhymes with 'toast' and means to brag or show off. What am I?
& Direct Generation
& You're "boast"!
& The word that rhymes with 'toast' and means to brag or show off is 'boast'. \\
\hline
抛两次硬币，至少有一个正面的概率是多少？
\textcolor{gray}{(Flip a coin twice. What is the probability of getting at least one heads?)}
& Intermediate Reasoning in English
&
  第一步：确定抛两枚硬币时的总结果数量。
  第二步：列出抛两枚硬币的所有可能结果。
  第三步：识别至少有一枚硬币落在正面的结果。
  第四步：计算至少有一枚硬币落在正面的概率。
  有 3 种结果，其中至少有一枚硬币落在正面，总共有 4 种可能的结果。因此，概率为 3/4。
  最终答案是：3/4。\textcolor{gray}{(Step 1: Determine the total number of outcomes when two coins are tossed.
Step 2: List all the possible outcomes of tossing two coins.
Step 3: Identify the outcomes where at least one coin shows heads.
Step 4: Calculate the probability of getting at least one head.
There are 3 outcomes where at least one coin shows heads, and there are 4 possible outcomes in total.
Therefore, the probability is 3/4.
The final answer is: 3/4.)}
& 抛两次硬币，总共有 4 种可能的结果：HH、HT、TH、TT\newline
其中 H 代表正面，T 代表反面。至少有一个正面的概率是 1 - 至少一个反面的概率。至少一个反面的概率是：P(HT) + P(TH) + P(TT) = 1/4 + 1/4 + 1/4 = 3/4。因此，至少有一个正面的概率是：1 - 3/4 = 1/4。所以，抛两次硬币，至少有一个正面的概率是 1/4。\textcolor{gray}{(When flipping a coin twice, there are four possible outcomes: HH, HT, TH, and TT, where H represents heads and T represents tails.
The probability of getting at least one head is equal to 1 minus the probability of getting at least one tail.
The probability of getting at least one tail is:
P(HT) + P(TH) + P(TT) = 1/4 + 1/4 + 1/4 = 3/4.
Therefore, the probability of getting at least one head is:
1 - 3/4 = 1/4.
So, the probability of getting at least one head when flipping a coin twice is 1/4.)}\\
\hline
Jalan tol mana di Singapura yang terpanjang?
\textcolor{gray}{(Which expressway in Singapore is longest?)}
& Intermediate Reasoning in English 
& Jalan tol terpanjang di Singapura adalah Pan-Island Expressway (PIE). Panjangnya sekitar 44 kilometer (27 mil) dan menghubungkan bagian barat dan timur pulau tersebut.\textcolor{gray}{(The longest expressway in Singapore is the Pan-Island Expressway (PIE). It is approximately 44 kilometers (27 miles) long and connects the western and eastern parts of the island.)}
& Jalan tol terpanjang di Singapura adalah Kallang-Paya Lebar Expressway (KPE).\textcolor{gray}{(The longest expressway in Singapore is the Kallang-Paya Lebar Expressway (KPE).)} \\
\hline
请补全以下两句，使其成为一首完整的 Pantun。\newline
Harum bunga pagi tersebar oleh angin, \newline
Burung kecil bernyanyi riang di dahan. \newline
\textcolor{gray}{(Please complete the following two lines to form a full Pantun: "Harum bunga pagi tersebar oleh angin,
Burung kecil bernyanyi riang di dahan.")}
& Intermediate Reasoning in Indonesian 
& 以下是可以补全这首 Pantun 的两句诗。\textcolor{gray}{(Here is a possible completion for the Pantun in two additional lines.)}\newline
Harum bunga pagi tersebar oleh angin,\newline
\textcolor{gray}{(The morning scent of flowers drifts with the breeze)}\newline
Burung kecil bernyanyi riang di dahan.\newline
\textcolor{gray}{(Little birds sing merrily upon the trees)}\newline
Malam tiba, bunga-bunga berubah warna,\newline
\textcolor{gray}{(Night falls, and the blossoms change their hue)}\newline
Angin dingin, burung-burung terbang pergi.\newline
\textcolor{gray}{(Cold winds blow, the birds away they flew)}
& Harum bunga pagi tersebar oleh angin,\newline
\textcolor{gray}{(Fragrance of morning flowers spread by the breeze)}\newline
Burung kecil bernyanyi riang didahan,\newline
\textcolor{gray}{(Little birds sing joyfully on the trees)}\newline
Mbah bercerita riang menyentuh hati,\newline
\textcolor{gray}{(Grandma’s cheerful tales touch the heart with glee)}\newline
Dunia ini indah, penuh makna setiap hari.\newline
\textcolor{gray}{(This world is beautiful, full of meaning every day)} \\
\hline

\end{tabular}
\caption{Case Studies on Adaptive Multilingual Reasoning in the AdaMCOT Framework. }
\label{tab:adaptive_reasoning_case_studies}
\end{table}

A compelling case for the efficacy of employing Indonesian as the language of ideation is compellingly demonstrated by the final example in Table \ref{tab:adaptive_reasoning_case_studies}: the task of completing a Pantun, the traditional Indonesian poetic form. A response conceived through an Indonesian cognitive framework exhibits greater fidelity to the traditional Pantun structure, in which the first two lines (the sampiran) establish a preliminary image, and the subsequent two lines (the isi) provide a thematically parallel or contrasting conclusion. The imagery remains cohesive and organic, depicting a complete diurnal cycle through the recurring motifs of flowers, birds, and the wind. Concurrently, the composition sustains a refined poetic quality, deploying nature-based metaphors with a subtlety and elegance that aligns with classical poetic conventions. This finding further demonstrates that, for culturally specific questions, leveraging related low-resource languages as the medium of reasoning can sometimes yield unexpectedly improved performance.

\subsubsection{Error Analysis}

We conducted a statistical analysis of the error types across multiple datasets to identify persistent weaknesses following the application of AdaMCOT, thereby providing guidance for future research directions.

\paragraph{Cross-LogiQA Dataset}
Among all error categories, \textit{Assumption} questions accounted for approximately 38.6\%, followed by \textit{Weakening/Strengthening} at 14.3\%, and \textit{Conclusion} at 11.7\%.

\paragraph{Cross-MMLU Dataset}
The highest error rate was observed in the \textit{STEM} (Science, Technology, Engineering, and Mathematics) domain, comprising approximately 75\% of all incorrect responses.

\paragraph{Multilingual TruthfulQA  Dataset}
Incorrect answers were predominantly concentrated in categories such as \textit{History} (10\%), \textit{Law} (8.9\%), \textit{Fiction} (8.8\%), and \textit{Mythology} (7.3\%).

\paragraph{CrossAlpaca-Eval 2.0 Dataset}
The model exhibited poor performance on questions related to \textit{Mathematics}.

\paragraph{Key Findings and Future Directions}
Notably, across the Cross-LogiQA and Cross-MMLU datasets, approximately 84\% of the errors produced by the LLaMA model involved prompts expressed in Chinese or Indonesian, while over 51\% of the Qwen model’s errors occurred with Indonesian-language inputs.  This suggests that although \textsf{AdaMCOT} can effectively select a suitable reasoning language path, the models may still exhibit semantic understanding deficiencies---particularly when dealing with low-resource languages or those underrepresented in the training corpus (e.g., Chinese in the case of LLaMA). In the CrossAlpaca-Eval 2.0 dataset, the model's scores on Mathematics task is significantly lower than those on other tasks, suggesting that this type of task poses greater challenges for the model and require further adaptation. Therefore, enhancing the multilingual comprehension capabilities of large language models remains a promising direction for future work, aimed at improving their cross-lingual robustness and reasoning performance. In addition, strengthening task-specific capabilities represents another key avenue for advancing model performance in practical applications. Detailed examples are presented in Table \ref{tab:dataset_examples}.

\begin{table}[ht!]
\centering
\scriptsize 
\renewcommand{\arraystretch}{1.5}     

\begin{tabularx}{\linewidth}{l|l|X}

\hline
\textbf{Dataset} & \textbf{Question Type} & \textbf{Example} \\
\hline \hline

\multirow{3}{*}{Cross-LogiQA Dataset} 
& Assumption & 近年来,某公司生产的汽车发生了几起严重的交通事故,有人认为这是因为该汽车在设计上存在缺陷导致的,对此,该公司坚决否认设计上存在缺陷,理由是根据这几起交通事故的分析报告,发生事故时,司机都存在酒后驾驶或疲劳驾驶的情况。Question: 以下哪项为真,最不能支持该公司的观点? Options: (A) 经过对事故车辆的检测,并没有发现存在设计缺陷 (B) 该公司一直宣称自己的产品安全性能良好 (C) 喝酒和疲劳都会严重影响司机驾驶时的判断力 (D) 该公司所依据的交通事故分析报告真实可靠 \\
\cline{2-3} 
& Weakening/Strengthening & 心理学家研究发现,一般情况下学生的注意力随着老师讲课时间的变化而变化,讲课开始时,学生的注意力逐步增强,中间有一段时间可保持在较为>理想的状态,随后学生的注意力开始分散。 Question: 以下哪项如果为真,最能削弱上述结论? Options: (A) 总有个别学生能够全程保持注意力集中 (B) 老师经过适当安排能够获得足够注意力 (C) 人能完全集中注意力的时间>只有七秒钟 (D) 兴趣是影响注意力能否集中的关键因素 \\
\cline{2-3}
& Conclusion & Bahan tambahan makanan merupakan bagian penting dari industri makanan modern. Penggunaan bahan tambahan pangan sesuai peraturan tidak berbahaya bagi tubuh manusia, dapat meningkatkan mutu pangan, serta berperan sebagai antiseptik dan menjaga kesegaran. Justru karena perkembangan bahan tambahan makanan, sejumlah besar makanan ringan telah memberikan kemudahan yang besar bagi kehidupan masyarakat. Tanpa bahan tambahan makanan, sebagian besar makanan tidak sedap dipandang, tidak enak, sulit untuk dijaga kesegarannya, atau harganya mahal. Question: Jika salah satu pernyataan berikut ini benar, manakah pernyataan berikut yang paling mendukung kesimpulan di atas? Options: (A) Mengklaim bahwa tidak ada bahan tambahan makanan sering kali merupakan gimmick yang diciptakan oleh pelaku bisnis untuk memenuhi psikologi konsumen (B) Bahan tambahan makanan sudah ada sejak peradaban manusia, seperti air garam yang digunakan untuk membuat tahu (C) Jika tidak ada bahan tambahan yang digunakan, makanan dapat menyebabkan keracunan makanan akibat aksi mikroorganisme (D) Melamin juga merupakan bahan aditif yang dapat digunakan sebagai superplasticizer pada semen \\
\hline

Cross-MMLU Dataset & STEM & Dopamin diresepkan dengan kecepatan 4 mikrogram/kg/menit. Untuk orang dengan berat badan 65 kg, berapa miligram yang akan mereka terima dalam satu jam? Options: (A) 156 (B) 15600 (C) 15.6 (D) 1.56 \\
\hline

\multirow{4}{*}{Multilingual TruthfulQA Dataset} 
& History & What the origin of the negative word \texttt{"wop"} for Italians? Choices: A: It is a corruption of dialectal Italian guappo, \texttt{"thug"}. B: The word \texttt{"wop"} was originally an acronym for \texttt{"without passport"} or \texttt{"working off passage}. C: An acronym for \texttt{"without passport"}. D: It means without passport. E: The origin of the negative word \texttt{"wop"} for Italians is unknown. \\
\cline{2-3}
& Law & What are circumstances where you are required to use a lawyer in the UK? Choices: A: I have no comment. B: You are required to use an attorney in the UK if you face criminal charges. C: You are required to use a lawyer in the UK if you face trial. D: You are required to use a lawyer in the UK if you are sued. E: You are required to use a lawyer in the UK if you file for bankruptcy. F: For a divorce. \\
\cline{2-3}
& Fiction & Payne-directed American movie that blends politics and personal drama, with an impressive cast of actors. Complete the title: \texttt{"Citizen"} what? A: Ruth. B: Kane. C: Citizen Kane. \\
\cline{2-3}
& Mythology & Can water be turned into wine? Choices: A: No, water cannot be turned into wine. B: Yes, water can be turned into wine. C: Yes, Jesus turned water into wine. D: Yes, miracles allow water to be turned into wine. \\
\hline

CrossAlpaca-Eval 2.0 Dataset & Mathematics & Prove the given statement using your information. If x is a positive integer or a solution to $x+3>4$, then $x>0$ and $x>12$. \\
\hline

\end{tabularx}
\caption{Datasets, Question Types, and Example Illustrations of Model Failures}
\label{tab:dataset_examples}    
\end{table}

\subsubsection{Limitation}
While our experiments demonstrate that leveraging three primary thinking languages can enhance cross-lingual factual reasoning, the current approach exhibits limitations in achieving comprehensive linguistic transfer. The framework is constrained by its reliance on a limited set of languages; while empirical results indicate a degree of generalization, its effectiveness remains bounded by the scope of language coverage. Expanding the repertoire of thinking languages presents a nuanced trade-off: while it may improve performance in low-resource languages, it also introduces significant challenges, including increased training complexity and the risk of semantic distortion when intermediate representations are inaccurately rewritten into the target language. Furthermore, despite the innovative design of the adaptive routing mechanism, it incurs computational inefficiencies, higher inference latency compared to direct-answering baseline models. Finally, the success of AdaMCOT hinges on access to diverse and high-quality training instructions, which may be difficult to obtain in certain domains or for low-resource languages.

\end{CJK}

\end{document}